\documentclass[letterpaper]{article} 
\usepackage{aaai25}  
\usepackage{times}  
\usepackage{helvet}  
\usepackage{courier}  
\usepackage[hyphens]{url}  
\usepackage{graphicx} 
\usepackage{natbib}
\usepackage[many]{tcolorbox}
\usepackage{tikz}
\usepackage{enumitem}
\usepackage{tabularx}

\usepackage{graphicx} 
\usepackage[utf8]{inputenc} 
\usepackage[T1]{fontenc}    
\usepackage{adjustbox}
\usepackage{url}            
\usepackage{booktabs}       
\usepackage{amsfonts}       
\usepackage{nicefrac}       
\usepackage{microtype}      
\usepackage{xcolor}

\definecolor{azure}{rgb}{0.0, 0.5, 1.0}
\definecolor{lightgreen}{RGB}{142, 207, 201}
\definecolor{lightblue}{RGB}{207,234,241}

\usepackage{amsmath}

\usepackage{appendix} 
\usepackage{enumitem}
\usepackage{longtable}
\usepackage{url}
\usepackage{xltabular}
\usepackage{colortbl}
\usepackage{subcaption}

\usepackage{caption} 
\usepackage{algorithm}
\usepackage{algorithmic}

\usepackage{newfloat}
\usepackage{listings}
\DeclareCaptionStyle{ruled}{labelfont=normalfont,labelsep=colon,strut=off} 
\floatstyle{ruled}
\newfloat{listing}{tb}{lst}{}
\floatname{listing}{Listing}
\title{UniAutoML: A Human-Centered Framework for Unified Discriminative and Generative AutoML with Large Language Models}
\author {
    Jiayi Guo\textsuperscript{\rm 1, \rm 2},
    Zan Chen\textsuperscript{\rm 3},
    Yingrui Ji\textsuperscript{\rm 4, \rm 5},
    Liyun Zhang\textsuperscript{\rm 6},
    Daqin Luo\textsuperscript{\rm 7},
    Zhigang Li\textsuperscript{\rm 8},
    Yiqin Shen\textsuperscript{\rm 9, *}
}
\affiliations {
    \textsuperscript{\rm 1} Glasgow College, University of Electronic Science and Technology of China, Chengdu, China\\
    \textsuperscript{\rm 2} School of Engineering, University of Glasgow, Glasgow, Scotland\\
    \textsuperscript{\rm 3} Toursun Synbio, Shanghai, China\\
    \textsuperscript{\rm 4} Aerospace Information Research Institute, Chinese Academy of Sciences, Beijing, China \\
    \textsuperscript{\rm 5} School of Electronic, Electrical and Communication Engineering, University of Chinese Academy of Sciences, Beijing, China \\
    \textsuperscript{\rm 6} School of Information and Software Engineering, University of Electronic Science and Technology Of China, Chengdu, China\\
    \textsuperscript{\rm 7}Research Institute, Ubtech Robotics Corp, Shengzhen, China\\
    \textsuperscript{\rm 8} Qiyuan Lab, Beijing, China\\
     \textsuperscript{\rm 9} Department of Computer Science, Johns Hopkins University, MD, USA\\
     * Corresponding author: yshen92@jhu.edu
}

\usepackage{bibentry}

\begin{document}

\maketitle

\begin{abstract}
Automated Machine Learning (AutoML) can streamline the complexities of machine learning pipelines.
However, existing AutoML frameworks focus on discriminative tasks leaving generative tasks unexplored. 
Moreover, existing AutoML frameworks often lack interpretability and user engagement during the model development and training process, due to the absence of human-centered designs. 
%
%
To address these challenges, we propose UniAutoML, a human-centered AutoML framework that leverages Large Language Models (LLMs) to unify the automation of both discriminative (\textit{e}.\textit{g}., Transformers and CNNs for classification or regression) and generative tasks (\textit{e}.\textit{g}., fine-tuning diffusion models or LLMs). 
UniAutoML enables natural language interactions between users and the framework to provide real-time feedback and progress updates, thus enhancing the interpretability, transparency, and user control, where users can seamlessly modify or refine the model training in response to evolving needs. 
To ensure the safety and reliability of LLM-generated content, UniAutoML incorporates a safety guard-line that filter inputs and censor outputs. 
We evaluated UniAutoML's performance and usability through experiments conducted on eight diverse datasets and user studies involving 25 participants. 
Our results demonstrate that UniAutoML not only achieves higher performance but also improves user control and trust in the AutoML process, which makes ML more accessible and intuitive for a broader audience.
The code for UniAutoML is available at: \url{https://github.com/richarf22/UniAutoML}.
\end{abstract}

\section{Introduction}
Automated Machine Learning (AutoML) has emerged as a user-friendly solution designed to simplify the complex process of building and training machine learning (ML) models \cite{ref1}.
By automating model selection, hyperparameter tuning, and code generation, AutoML makes ML more accessible to individuals without ML expertise. 
This automation not only accelerates model development but also reduces the cognitive load for both novice and experienced users by simplifying complex workflows and minimizing repetitive tasks.
However, existing AutoML frameworks, such as AutoGluon \cite{erickson2020autogluon} and Auto-sklearn \cite{feurer-arxiv20a}, predominantly focus on discriminative models, leaving a significant gap in their applicability to generative models—an area of growing importance, especially with the rise of diffusion models \cite{ho2020denoising} and Large Language Models (LLMs). This limitation is particularly challenging for non-experts who might need to engage with generative modeling tasks but find current AutoML solutions inadequate.
Furthermore, traditional AutoML frameworks often provide limited interpretability and transparency, offering minimal explanations for the steps involved and lacking real-time feedback during the AutoML process. 
This ``black box'' nature can diminish user trust and hinder the adoption of AutoML tools, especially among those who seek to understand and interact with the models they use. 
The inability to easily modify or fine-tune selected models further exacerbates this issue, underscoring the need for a more human-centered approach to AutoML that bridges the gap between discriminative and generative models while enhancing user engagement and trust.

To address these limitations, we introduce UniAutoML, a novel framework that unifies AutoML for both discriminative and generative tasks. 
UniAutoML not only supports traditional discriminative models such as Transformers and Convolutional Neural Networks (CNNs) for tasks like classification and regression but also extends its capabilities to advanced generative models, including diffusion models and large language models (LLMs). A key feature of UniAutoML is its automated fine-tuning process, enabling users to train these complex models effortlessly through natural language commands, eliminating the need for manual coding. 
For instance, users can fine-tune diffusion models by simply providing datasets, while LLMs can be fine-tuned using the XTuner \cite{2023xtuner} tool by specifying a configuration name or address.
This approach significantly lowers the barriers for non-experts, allowing them to engage with advanced generative and discriminative models without requiring deep technical expertise.
Another innovation in UniAutoML is its human-centered design, featuring a conversational user interface (CUI) that allows users to interact with the model through natural language. 
UniAutoML eliminates the need for users to have prior knowledge of Python or specific AutoML tools, thereby broadening the accessibility of AutoML.
The CUI offers an intuitive and interactive experience, guiding users step-by-step through the AutoML process while providing real-time updates on model selection, training progress, and performance metrics.
This continuous feedback loop not only enhances transparency but also empowers users to make informed decisions, such as adjusting model parameters or terminating underperforming trials mid-process.
Additionally, UniAutoML leverages the vast repository of pre-trained models available through HuggingFace, significantly expanding the range of models accessible to users compared to traditional frameworks like AutoKeras and Auto-sklearn, which are confined to their respective pre-trained modules. 
It not only increases the versatility of UniAutoML but also ensures that users have access to the latest and most suitable models for their tasks with minor efforts on maintenance.
To realize the potential of UniAutoML, we employ LLMs as the core component of the framework. 
However, recognizing the potential risks associated with LLM-generated content, we design a safety guard-line that filters user inputs and censors outputs, ensuring that UniAutoML remains focused on AutoML tasks and mitigating potential security or ethical concerns.

\begin{figure*}[t!]
  \centering
  \includegraphics[width=\linewidth]{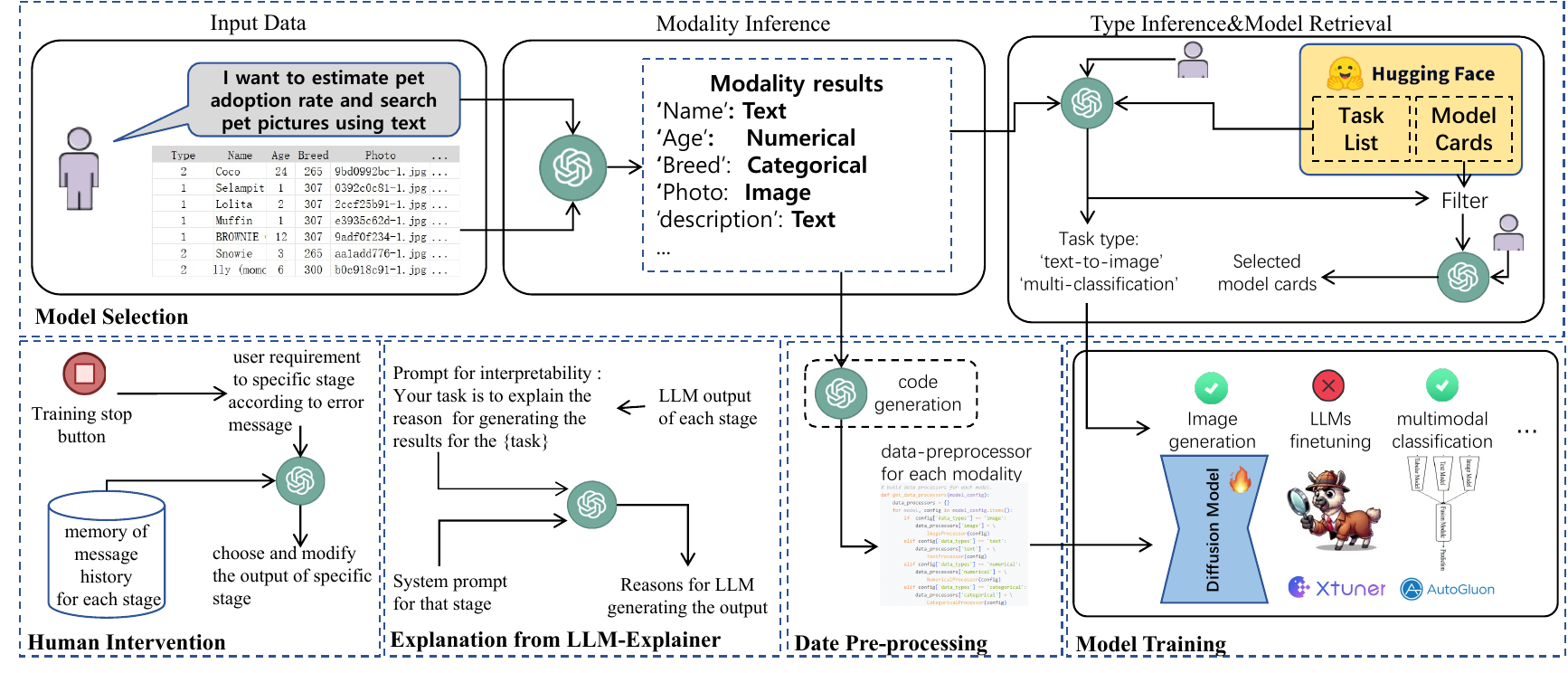}
  \caption{Overall framework of the proposed UniAutoML. }
  \label{fig1}
\end{figure*}

The major contributions are five-fold.
Firstly, we propose the first AutoML framework that unifies both discriminative and generative tasks, including automated fine-tuning for diffusion models and LLMs. 
This integration reduces the learning curve for users working with these advanced models, making complex ML tasks more accessible to non-experts.
Secondly, we introduce a human-centered design to enhance both user control and interpretability. 
It not only provides detailed AutoML outputs and progress information but also enables users to actively participate in the AutoML process. 
Users can modify model parameters, terminate underperforming trials, and make informed decisions in real-time, fostering a deeper understanding and more effective control over the entire ML pipeline.
Thirdly, UniAutoML features connectivity with HuggingFace, allowing it to retrieve and utilize a wide range of pre-trained models, which thus eliminates the need for local maintenance of model libraries and significantly enhances the scalability and versatility of AutoML. 
Users can access the latest and most suitable models for their tasks with minimal effort, keeping pace with rapid advancements in the field.
Forth, we implement a safety guard-line mechanism that addresses potential safety and ethical concerns associated with LLM use.
It carefully inspects both user inputs and LLM outputs, maintaining high standards of security and ethicality within our UniAutoML framework.
Finally, we conduct a comprehensive evaluation of UniAutoML through both quantitative experiments and user studies.

\section{Methods}
\label{gen_inst}
\paragraph{Overview of UniAutoML}
UniAutoML is an innovative human-centered AutoML framework that leverages LLMs to unify the automation of both discriminative and generative machine learning tasks, as illustrated in Fig.~\ref{fig1}.
Users initiate by providing task requirements and datasets descriptions through plain natural language. 
The model selection module, implemented as LLMs, then identifies the most suitable pre-trained models from HuggingFace by comparing model features with task requirements and dataset characteristics, supporting both discriminative and generative tasks. 
For model configuration, UniAutoML acquires corresponding settings from HuggingFace and defines the hyperparameter search space.
To ensure efficient handling of diverse data modalities, the data pre-processing module generates code for data loading and preparation.
The model training module then constructs the pipeline for model training, incorporating techniques like Low-Rank Adaptation (LoRA) \cite{hu2021lora} for fine-tuning generative models when necessary.
Throughout this process, an LLM-Explainer module provides real-time interpretations of each decision and output, enhancing user understanding and control.
Additionally, a safety guard-line mechanism is implemented to filter both user inputs and LLM outputs, preventing the generation or propagation of harmful or irrelevant information.

\paragraph{Unified AutoML for Discriminative and Generative Tasks}
UniAutoML seamlessly integrates support for both discriminative and generative tasks within a single framework.
Importantly, the pre-processing, model configuration selection, and safety guard-line modules are shared across both discriminative and generative tasks, ensuring a consistent and efficient workflow regardless of the task type.
The key differences lie in the model selection and model training modules, which are tailored to the specific requirements of discriminative and generative tasks.
For discriminative tasks, UniAutoML the model selection builds upon the structure of AutoGluon \cite{erickson2020autogluon}, supporting a wide array of models suitable for classification, regression, and other predictive tasks across multiple data modalities, including image, text, and tabular data. 
Specifically, the model selection module in UniAutoML is powered by LLM to ensure that the most appropriate pre-trained discriminative models are chosen based on the specific requirements of the task and the characteristics of the input data.
In the realm of generative tasks, UniAutoML introduces automation for fine-tuning both diffusion models and LLMs.
For diffusion models, users simply provide the dataset address and express their intention to use a generative model (\textit{e}.\textit{g}., diffusion model) through natural language.
UniAutoML then selects the diffusion model with the model selection module, and the model training module employs LoRA to fine-tune the model on the provided dataset. 
Once training is complete, users can generate images by inputting prompts through the CUI again, leveraging the capabilities of the fine-tuned model.
LLM fine-tuning in UniAutoML is facilitated through seamless integration with XTuner \cite{2023xtuner}. 
Users need only specify the name or path of the LLM configuration they intend to fine-tune, along with their desired storage path for the resulting model.

\paragraph{Model Selection Module}
This module in UniAutoML leverages LLM to comprehend and categorize task requirements based on the user query. 
The process begins by determining whether the task is generative or discriminative, along with identifying the relevant data modalities. 
In this step, the LLM acts as both a binary classifier for task type and an information extractor for modalities, represented as $t_c = \texttt{LLM}(p_c; q)$ and $\text{Modality}_d = \texttt{LLM}(p_m; q)$,
where $t_c$ is the task category (generative or discriminative), $\text{Modality}_d$ is the set of identified data modalities, $p_c$ and $p_m$ are prompts for classification and modality extraction respectively, and $q$ is the user query.
Afterwards, a more specific task type such as image classification, text generation, or tabular regression is deduced, represented as: 
\begin{equation}
\nonumber
s_t = \arg\max_{t \in T} \Big(\texttt{sim}(\texttt{Emb}(q \oplus t_c \oplus \text{Modality}_d), \texttt{Emb}(t)\Big), 
\end{equation}
where $s_t$ is the determined specific task type, $T$ is the set of all task types, $\texttt{sim}$ is a similarity function (\textit{i}.\textit{e}., cosine similarity) between the embeddings, $\texttt{Emb}$ refers to the embedding function that converts text into dense vector representations, and $\oplus$ denotes concatenation. 
The model cards from HuggingFace, containing metadata for both generative and discriminative models, are encoded in the same vector format to create separate vector databases for each task category.
Let $D_g$ and $D_d$ represent the vector databases for generative and discriminative models respectively: $D_g = {\texttt{Emb}(m) | m \in M_g}$ and $D_d = {\texttt{Emb}(m) | m \in M_d}]$, where $M_g$ and $M_d$ are the sets of generative and discriminative models. 
This separation ensures that the model selection is tailored to the specific requirements of generative or discriminative tasks.
Next, the selected task type $s_t$ is compared against the appropriate vector database of models to identify an informed shortlist of 10 candidate models as follows:
\begin{equation}
\nonumber
M_c = \texttt{top}_k\Big({\texttt{sim}(\texttt{Emb}(s_t), \texttt{Emb}(m)) | m \in D}, k=10\Big)
\end{equation}
where $M_c$ is the set of candidate models, $D$ is either $D_g$ or $D_d$ depending on $t_c$, and $\texttt{top}_k$ selects the $k$ models with the highest similarity scores.
Finally, these shortlisted models are further evaluated against the specific task requirements to identify the most suitable model:
\begin{equation}
m_b = \arg\max_{m \in M_c} \Big( \texttt{sim}(\texttt{Emb}(q), \texttt{Emb}(m)\Big),
\end{equation}
where $m_b$ is the best model selected based on the comparison between the candidate models $M_c$ and user query $q$.

\paragraph{Data Pre-processing and Model Training Modules}
To ensure efficient handling of diverse data modalities, UniAutoML's data pre-processing module generates code for data loading and preparation. This module adapts to the specific requirements of both discriminative and generative tasks, ensuring that the data is appropriately formatted for the selected model. The model training module then constructs the pipeline for model training, incorporating techniques like LoRA for fine-tuning generative models when necessary.
For discriminative tasks, it builds upon AutoGluon, while for generative tasks, it integrates specialized fine-tuning approaches for diffusion models and LLMs.

\paragraph{Enhanced Interpretability and User Interaction in UniAutoML}
At the core of our interpretation mechanism lies an LLM-Explainer module, that is designed to elucidate the rationale behind each decision of UniAutoML as well as the output into plain and user understanable language.
For each stage, once an output is yielded, it is transmitted to the LLM-Explainer module, which provides an explanation, represented as:
\begin{equation}
\text{Exp}_r = \texttt{LLM-Explainer}(p_e; r \oplus q \oplus c),
\end{equation}
where $\text{Exp}_r$ is the explanation for the result, $p_e$ is the system prompt for the LLM-Explainer, $r$ is the output result, and $c$ is the context information \textit{e}.\textit{g}., model card, modality information. 
For instance, during the model selection stage, after determining the most suitable model for the given task, the model card, user instructions, and modality information are conveyed to the LLM-Explainer as follows:
\begin{equation}
\nonumber
\text{Exp}_{m_b} = \texttt{LLM-Explainer}(p_e; m_b \oplus q \oplus t_c \oplus \text{Modality}_d),
\end{equation}
where $\text{Exp}_{m_b}$ is the explanation for the selected model $m_b$.
This design assists users in grasping the decision-making mechanisms of UniAutoML but also boosts their confidence in the validity of the chosen model. 
Moreover, the LLM-Explainer is capable of interpreting error information, providing a comprehensive breakdown with probable causes and potential resolutions:
\begin{equation}
\text{Exp}_{err} = \texttt{LLM-Explainer}(p_{err}; err \oplus q \oplus c),
\end{equation}
where $\text{Exp}_{err}$ is the explanation to the error message, $p_{err}$ is the prompt for error interpretation, and $err$ is the error information.
Users are prompted at the end of each stage with information about the completed phase as well as the upcoming stage. 
They are also given the opportunity to provide additional instructions for the next stage:
\begin{equation}
I_{next} = \texttt{LLM}(p_n; s_c \oplus s_n \oplus q),
\end{equation}
where $I_{next}$ is the information about the next stage, $p_n$ is the prompt for next stage information, $s_c$ is the current stage, $s_n$ is the next stage, and $q$ is the user query.
Furthermore, we have incorporated a progress indication tool, particularly during the model training stage. 
This live tracking mechanism provides users with an estimated completion time and a sense of temporal orientation.

\paragraph{Enhance User Control with Human-Centered Design}
The interaction of UniAutoML with user introduces several advancements to improve user control and engagement. 
Firstly, addressing the need for adaptability, UniAutoML introduces an interactive override feature where users can manually interrupt the workflow if intermediate results don't meet their expectations based on the LLM-Explainer module or if additional guidance is necessary. 
Specifically, UniAutoML is engineered to recognize the context of these user interventions precisely, discerning the specific stage they pertain to, allowing the model to retrace its steps and recommence operations from the implicated juncture, seamlessly integrating new directives.
This design maintains the delicate balance between automation efficiency and user discretion, allowing alterations to the automatic course without compromising the integrity of the entire process, whether for discriminative or generative tasks.
Moreover, leveraging the LLM-Explainer module, UniAutoML provides real-time explanations and feedback on each decision made during the AutoML process, where users can query the system about its choices, request clarifications, or suggest modifications at any point. 
This continuous dialogue ensures that users understand the rationale behind each step and can steer the process towards their specific requirements. 
For generative tasks, such as fine-tuning diffusion models or LLMs, this feature is particularly valuable as it allows users to refine the training process based on intermediate results or specific domain knowledge.

\paragraph{Safety Guard-Line Module}
LLM can present potential safety and ethical concerns. 
To address these issues and ensure the responsible use of LLMs within UniAutoML, we have implemented a safety guard-line module, which acts as a safeguard to monitor and filter both user inputs and LLM outputs throughout the AutoML pipeline.
The safety guard-line module operates in two stages. 
First, it scrutinizes user instructions to filter out irrelevant or potentially harmful information to ensure that only appropriate and task-relevant inputs are processed by the LLM, formulated as
$I_{filtered} = \texttt{LLM}(p_{filter}; I_{user})$
where $I_{filtered}$ is the filtered input, $p_{filter}$ is the system prompt for input filtering, and $I_{user}$ is the original user input.
Following the processing of filtered inputs, this module then examines the LLM's output for any content that may compromise user safety or raise ethical concerns. 
If such content is detected, the module prompts the LLM to revise its output. 
This iterative refinement process can be represented as
$O_{safe} = \texttt{LLM}(p_{revise}; O_{initial} \oplus C_{critique})$
where $O_{safe}$ is the safe, revised output, $p_{revise}$ is the system prompt for output revision, $O_{initial}$ is the initial LLM output, and $C_{critique}$ is the critique of the initial output.

\section{Experiments}
\paragraph{Implementation Details}
We conducted our experiments using two NVIDIA A6000 GPUs with CUDA 11.7. 
The software environment consisted of Python 3.8 and PyTorch 2.0.1. 
For the LLM component, we utilized GPT-4o \cite{openai_gpt4}.
%
To ensure consistency and reproducibility, we set the `Temperature' parameter, which controls the model's creativity, to 0. 
%
%
For a detailed description of the prompts used at each stage of our framework, please refer to Appendix.

\subsection{Quantitative Evaluation on Discriminative Tasks}
Our quantitative evaluation assesses the performance of UniAutoML in comparison to other AutoML frameworks. 
We focused on two aspects, namely the model selection module and the model training module. 
It's important to note that in this evaluation, we concentrate solely on discriminative tasks, because there are currently no existing AutoML frameworks for generative models. 
Consequently, the evaluation of UniAutoML's performance on generative AutoML tasks will be addressed from a human-centric perspective in our user study.

\paragraph{Baselines}
We compared UniAutoML against two prominent AutoML frameworks, AutoGluon \cite{erickson2020autogluon} (version 1.0.0) and AutoKeras \cite{JMLR:v24:20-1355} (version 2.0.0). These baselines were chosen to represent different approaches within the AutoML landscape. AutoKeras primarily employs neural architecture search (NAS), while AutoGluon focuses on hyperparameter optimization (HPO). 
To ensure a fair comparison, we configured these baselines with equivalent dataset addresses, training quality parameters, and epoch settings as used in UniAutoML.

\begin{table}[htbp!]
\caption{Overview of datasets used for evaluation.}
\label{da}
\centering
\resizebox{\linewidth}{!}{
\begin{tabular}{lrrlll}
\toprule
Dataset & Train Samples & Test Samples & Task Type & Metric & Description \\
\midrule
SPMG    & 5,000         & 1,000        & Retrieval   & AUC    & Product matching \\
PPCD    & 8,920         & 993          & Regression  & RMSE   & Pet popularity prediction \\
PARA    & 25,398        & 2,822        & Regression  & RMSE   & Image aesthetics assessment \\
CH-SIMS & 2,052         & 228          & Multiclass  & AUC    & Sentiment classification \\
LMSYS   & 57,477        & -            & Multiclass  & AUC    & Language model evaluation \\
TMLFDD  & 891           & 418          & Binary      & AUC    & Survival prediction \\
PAPD    & 13,493        & 1,499        & Multiclass  & AUC    & Pet adoption speed prediction \\
MMSDD   & 17,833        & 1,981        & Binary      & AUC    & Sarcasm detection \\
\bottomrule
\end{tabular}
}
\end{table}

\paragraph{Datasets}
To evaluate UniAutoML's performance across various tasks and data modalities, we selected a diverse set of datasets, primarily sourced from Kaggle\footnote{\url{https://www.kaggle.com/}}. 
Table \ref{da} provides an overview of the eight multimodal datasets used in our experiments, detailing their size, task type, and evaluation metrics.
The datasets encompass a range of ML tasks, including classification, regression, and retrieval. 
For classification tasks, we employed the Multi-Modal Sarcasm Detection (MMSDD) and Titanic Machine Learning from Disaster (TMLFDD) datasets for binary classification, while the PetFinder.my-Adoption Prediction (PAPD), LMSYS, and CH-SIMS datasets were used for multiclass classification. 
To evaluate regression capabilities, we utilized the PARA dataset for image aesthetics assessment and the PetFinder.my-Pawpularity Contest (PPCD) dataset for predicting pet popularity. 
The Shopee-Price Match Guarantee (SPMG) dataset was chosen to assess UniAutoML's performance on retrieval tasks.

\begin{table}[htbp!]
\caption{
Performance comparison of the model selection module between AutoGluon and UniAutoML.
Results marked with `*' indicate manual correction necessary, otherwise model fails to converge.
Values represent in the form mean $\pm$ standard deviation.}
\label{model_selection}
\centering
\begin{tabular}{lcccc}
\toprule
Dataset & \textbf{AutoGluon} & \textbf{UniAutoML} \\
\midrule
SPMG ($\uparrow$) & 0.985{\scriptsize$\pm$0.004} & 0.987{\scriptsize$\pm$0.003} \\
PPCD ($\downarrow$) & 17.82{\scriptsize$\pm$0.296} & 17.74{\scriptsize$\pm$0.388} \\
PARA ($\downarrow$) & 0.569{\scriptsize$\pm$0.011} & 0.573{\scriptsize$\pm$0.019} \\
CH-SIMS ($\uparrow$) & 0.540{\scriptsize$\pm$0.025}* & 0.583{\scriptsize$\pm$0.031} \\
LMSYS ($\uparrow$) & 1.000{\scriptsize$\pm$0.000} & 1.000{\scriptsize$\pm$0.000} \\
TMLFDD ($\uparrow$) & 1.000{\scriptsize$\pm$0.000} & 1.000{\scriptsize$\pm$0.000} \\
PAPD ($\uparrow$) & 0.413{\scriptsize$\pm$0.009} & 0.414{\scriptsize$\pm$0.012} \\
MMSDD ($\uparrow$) & 0.953{\scriptsize$\pm$0.005} & 0.965{\scriptsize$\pm$0.002} \\
\bottomrule
\end{tabular}
\end{table}

\begin{figure}[htbp]
    \centering
    \includegraphics[width=\columnwidth]{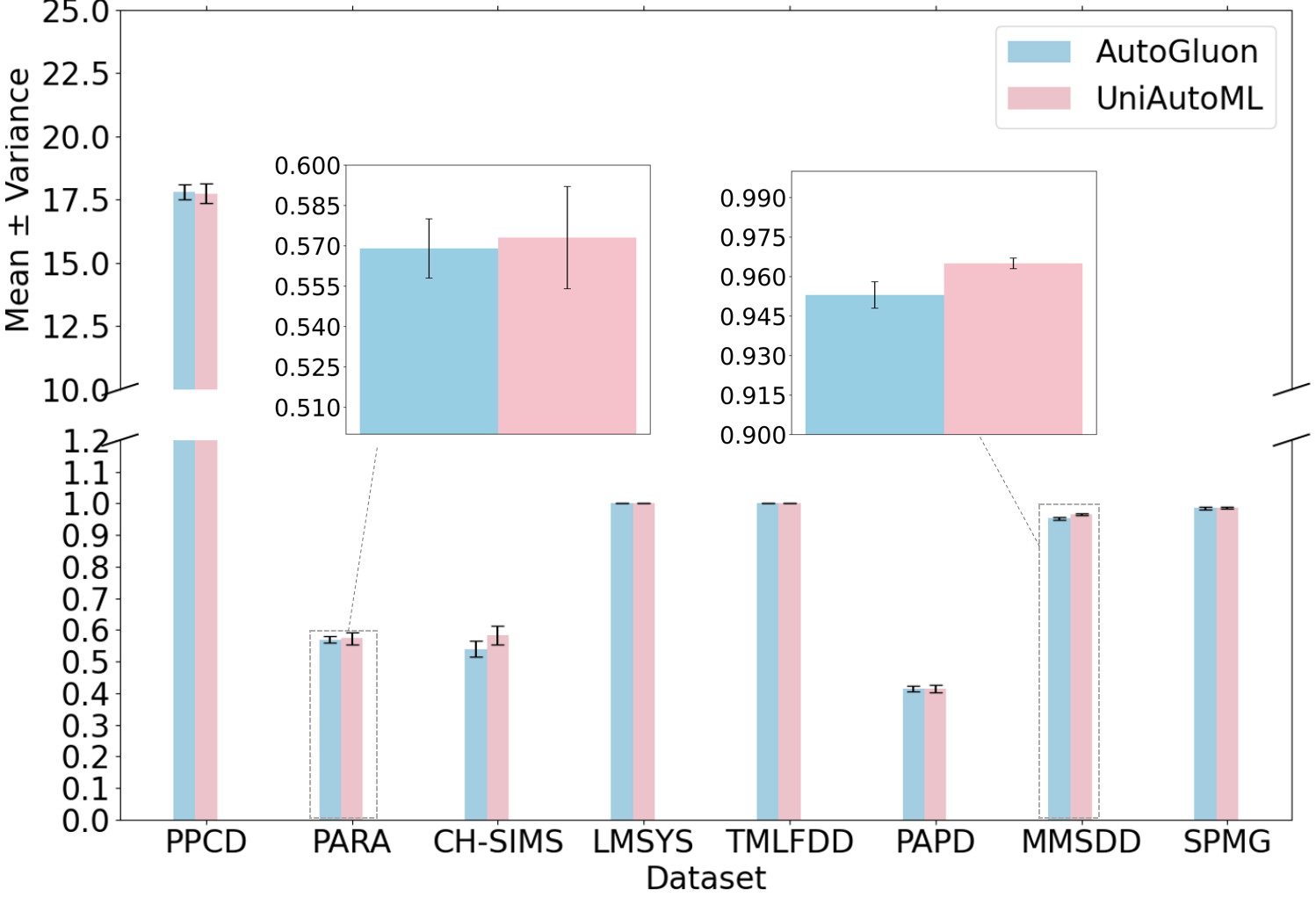}
     \caption{Performance comparison of the model selection module between AutoGluon and UniAutoML. }
  \label{fig2:combined}
\end{figure}

\paragraph{Performance on Model Selection Module}
We compared UniAutoML's model selection capabilities with those of AutoGluon, excluding AutoKeras from this comparison due to its reliance on manually defined column modalities rather than automated model selection. 
In our evaluation, we isolated the performance of the model selection module by keeping all other components, including model training, identical between AutoGluon and UniAutoML.  
Table \ref{model_selection} and Fig.~\ref{fig2:combined} present that UniAutoML demonstrates superior or comparable performance in most cases.
For the SPMG dataset, UniAutoML slightly outperforms AutoGluon with a score of $0.987$ compared to $0.985$. 
In the PPCD dataset, where lower scores are better, UniAutoML achieves a lower RMSE of $17.74$ versus AutoGluon's $17.82$. 
The CH-SIMS dataset shows an improvement with UniAutoML, scoring $0.583$ compared to AutoGluon's $0.540$, which required manual correction. 
For the PARA dataset, UniAutoML's performance ($0.573$) is marginally lower than AutoGluon's ($0.569$), though the difference is minimal. 
Both frameworks achieve perfect scores ($1.000$) for the LMSYS and TMLFDD datasets. 
The PAPD dataset shows a slight advantage for UniAutoML ($0.414$) over AutoGluon ($0.413$). 
Finally, for the MMSDD dataset, UniAutoML also outperforms AutoGluon with scores of $0.965$ and $0.953$, respectively.
These results highlight the advantage of UniAutoML's LLM-based approach, which can adapt to diverse data types without relying on predefined rules. 
The UniAutoML's ability to understand and process natural language enables it to make more nuanced decisions in model selection, particularly for complex, multimodal datasets.

\begin{table}[htbp]
\caption{Performance evaluation on model training module. 
Results marked with `*' indicate manual correction necessary, otherwise model fails to converge. 
`/' denotes that the task is not applicable with AutoKeras.
Values are presented as mean $\pm$ standard deviation.
}
\label{model_training}
\centering
\resizebox{\columnwidth}{!}{
\begin{tabular}{lcccc}
\toprule
\textbf{Dataset} & \textbf{AutoGluon w/HPO} & \textbf{UniAutoML} & \textbf{AutoGluon w/o HPO} & \textbf{AutoKeras} \\
\midrule
SPMG ($\uparrow$) & 0.986{\scriptsize$\pm$0.005} & 0.993{\scriptsize$\pm$0.004} & 0.990{\scriptsize$\pm$0.004} & / \\
PPCD ($\downarrow$) & 17.82{\scriptsize$\pm$0.296} & 17.44{\scriptsize$\pm$0.196} & 17.59{\scriptsize$\pm$0.230} & 23.20{\scriptsize$\pm$0.276} \\
PARA ($\downarrow$) & 0.569{\scriptsize$\pm$0.011} & 0.573{\scriptsize$\pm$0.019} & 0.562{\scriptsize$\pm$0.017} & 0.747{\scriptsize$\pm$0.019} \\
CH-SIMS ($\uparrow$) & 0.540{\scriptsize$\pm$0.025}* & 0.589{\scriptsize$\pm$0.027} & 0.561{\scriptsize$\pm$0.021}* & / \\
LMSYS ($\uparrow$) & 1.000{\scriptsize$\pm$0.000} & 1.000{\scriptsize$\pm$0.000} & 1.000{\scriptsize$\pm$0.000} & 1.000{\scriptsize$\pm$0.000} \\
TMLFDD ($\uparrow$) & 1.000{\scriptsize$\pm$0.000} & 1.000{\scriptsize$\pm$0.000} & 1.000{\scriptsize$\pm$0.000} & 1.000{\scriptsize$\pm$0.000} \\
PAPD ($\uparrow$) & 0.413{\scriptsize$\pm$0.009} & 0.445{\scriptsize$\pm$0.014} & 0.442{\scriptsize$\pm$0.009} & 0.383{\scriptsize$\pm$0.013} \\
MMSDD ($\uparrow$) & 0.953{\scriptsize$\pm$0.005} & 0.965{\scriptsize$\pm$0.002} & 0.922{\scriptsize$\pm$0.008} & / \\
\bottomrule
\end{tabular}
  }
\end{table}

\begin{figure}[htbp!]
    \centering
    \includegraphics[width=1\columnwidth]{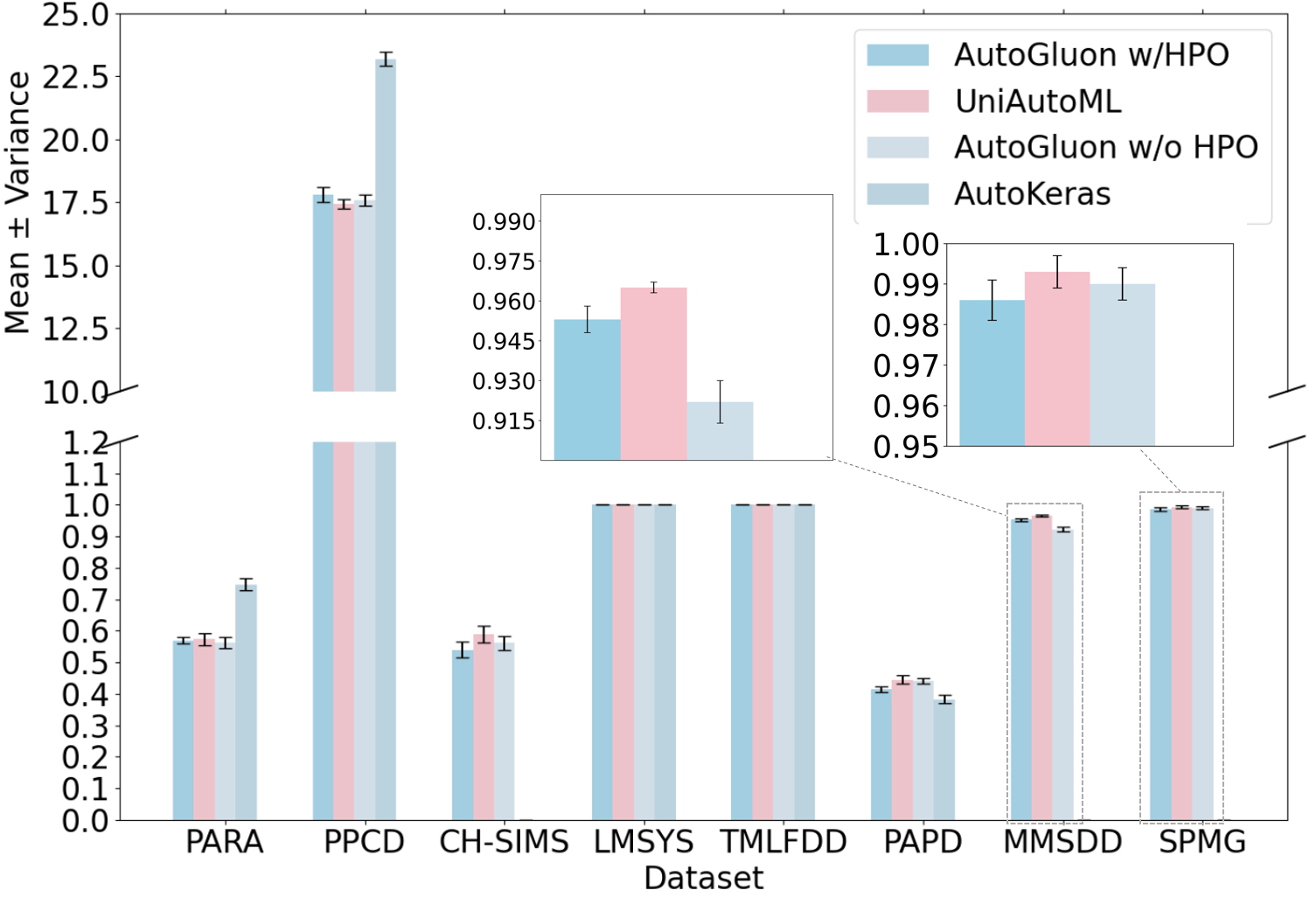}
     \caption{Performance comparison on model training module.} 
  \label{fig1:combined}
\end{figure}

\paragraph{Performance on Model Training Module}
We evaluated the effectiveness of model training, including hyperparameter optimization, across UniAutoML, AutoGluon (with and without HPO), and AutoKeras in Table \ref{model_training} and Fig.~\ref{fig1:combined}.
To ensure a fair comparison, we utilized the same model selection results for all frameworks and focused solely on the performance of their respective training modules.
For the SPMG dataset, UniAutoML achieves the highest score of $0.993$, outperforming both versions of AutoGluon. 
In the PPCD dataset, where lower scores are better, UniAutoML shows the best performance with an RMSE of $17.44$, followed closely by AutoGluon without HPO ($17.59$), while AutoKeras lags more ($23.20$). 
The CH-SIMS dataset highlights a substantial improvement with UniAutoML, scoring $0.589$ compared to AutoGluon's $0.540$ (with HPO) and $0.561$ (without HPO), both of which required manual correction. 
%
%
These results demonstrate UniAutoML's robust performance across diverse datasets, often surpassing or matching the capabilities of established AutoML frameworks in model training, particularly in complex, multimodal scenarios.
In summary, our quantitative evaluation demonstrates that UniAutoML achieves competitive, and in many cases superior, performance in both model selection and model training modules on discriminative tasks.

\subsection{User Study}
To comprehensively evaluate UniAutoML's impact on user experience and its effectiveness in both discriminative and generative AutoML tasks, we conducted an extensive user study. 
This study was an complement to the quantative evaluation for assessing UniAutoML's performance on generative tasks, as no existing AutoML frameworks currently support such tasks for quantitative comparison.
Our investigation focused on four key aspects, formulated as the following null hypotheses: UniAutoML does not reduce the number of attempts required to complete a task, decrease user workload, improve system usability, or reduce the time required for users to accomplish tasks. 
To test these hypotheses, we employed single-sided t-tests to evaluate the statistical significance of differences in system usability, number of attempts, workload, and task completion time between UniAutoML and traditional AutoML counterpart.





\begin{figure}[htbp]
    \centering
    \includegraphics[width=1\columnwidth]{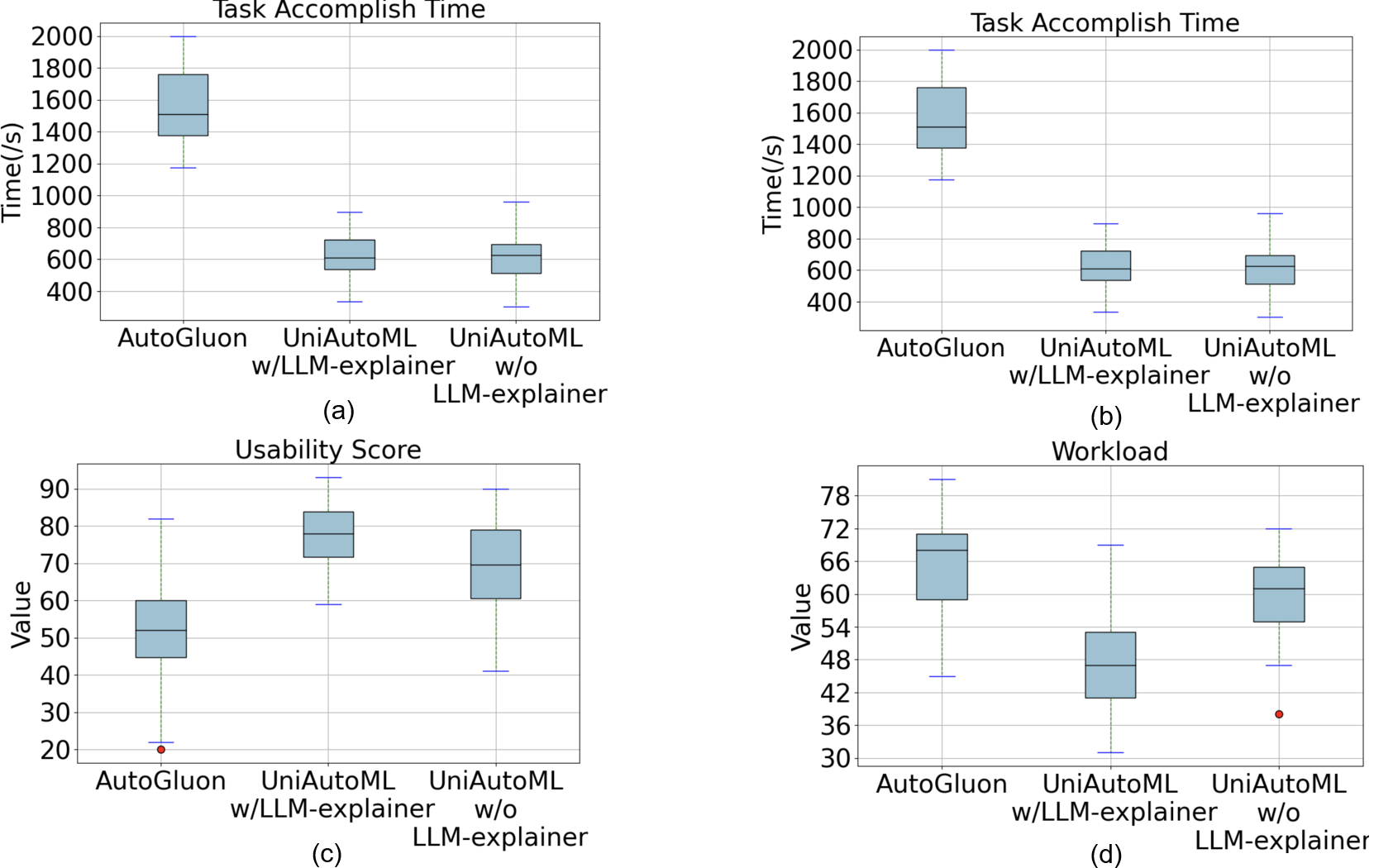}
    \caption{This figure has shown accomplishing time in a), attempt times in b), usability score in c) and workload in d) for AutoGluon, UniAutoML with interpretation and UniAutoML without interpretation.}
    \label{fig2}
\end{figure}

\paragraph{User Study Design}
\label{um}
The study was conducted in five stages, ensuring ethical considerations and diverse participant backgrounds.
We began by recruiting 25 volunteers from various fields, all potential users of AutoML frameworks. 
This diverse group included 15 Computer Science students, 5 teachers from non-computer science fields, 2 AI researchers, and 3 communication engineering teachers, allowing for a representative evaluation of UniAutoML.
Prior to the study, participants were briefed on the background, objectives, and procedures for UniAutoML and its counterpart. 
We emphasized their right to withdraw at any time without consequences, ensuring ethical compliance and participant comfort.
We then collected participant background information through a questionnaire.
The core of our study, illustrated in Fig. \ref{fig2}, involved dividing participants into three groups, each assigned to use one of three systems: AutoGluon, UniAutoML with LLM-Explainer, or UniAutoML without LLM-Explainer. 
Participants rotated through all three conditons over the course of the study.
During each session, participants were tasked with solving preset ML problems using their assigned AutoML method. 
These tasks included both discriminative problems (such as classification and regression) and generative tasks (such as fine-tuning diffusion models and LLMs). 
This approach allowed us to assess UniAutoML's unique capability in handling generative AutoML tasks, a feature not available in traditional AutoML frameworks.
After completing each task, participants filled out a detailed questionnaire (see Appendix) using Feishu Forms, providing quantitative and qualitative feedback on their experience.
This feedback covered aspects such as ease of use, perceived workload, and satisfaction with the system's performance on both discriminative and generative tasks.
By rotating participants through different coditions and collecting detailed feedback after each session, we were able to control for individual differences and gather comprehensive insights into UniAutoML's strengths and potential areas for improvement.

\paragraph{Results and Analysis }
The data collected for our four variables (number of attempts, user workload, system usability, and task completion time) is visualized in Fig.~\ref{fig2}. 
To validate our hypotheses, we conducted single-sided t-tests with a significance level of 0.05. 
For each hypothesis, we performed two separate t-tests with a significance level of 0.05: one comparing AutoGluon with UniAutoML with LLM-Explainer, and another comparing AutoGluon with UniAutoML without LLM-Explainer. 
As presented in Table \ref{hypo}, all null hypotheses were rejected, providing strong evidence that UniAutoML significantly improves user experience across all measured dimensions. 
These results demonstrate that the human-centered AutoML framework with LLM-Explainer has made it substantially easier for users to learn and control the AutoML process.
Notably, UniAutoML's performance was also impressive in generative AutoML tasks, where traditional frameworks like AutoGluon lack capabilities. 
Participants reported high satisfaction with UniAutoML's ability to handle tasks such as fine-tuning diffusion models and LLMs.
Additionally, many participants found the LLM-based interface intuitive, likely due to their prior experience with similar language models in other contexts.
As noted in Appendix, even AI researchers who were familiar with other AutoML tools found UniAutoML ``very interesting'' to use. 
This suggests that our UniAutoML not only lowers barriers for beginners but also offers engaging experiences for experienced practitioners.

\begin{figure}
    \centering
    \includegraphics[width=1\linewidth]{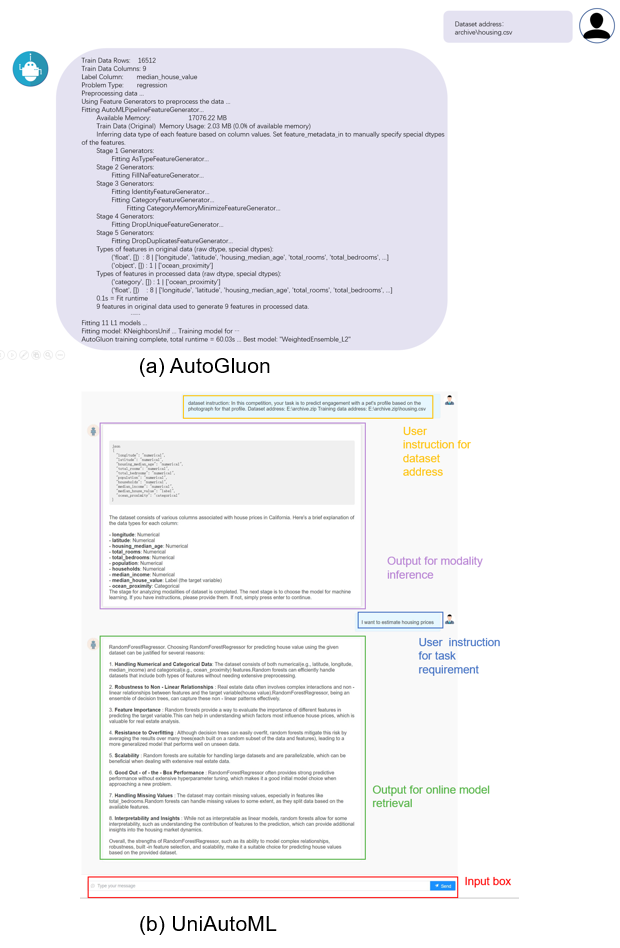}
    \caption{Comparative case study illustrating the general process flow of UniAutoML versus AutoGluon. UniAutoML's human-centered design features a conversational user interface, real-time explanations, and opportunities for user intervention at multiple stages.}
    \label{case}
\end{figure}

\begin{table}[htbp!]
  \caption{Statistical analysis of UniAutoML's impact on user experience via t-test comparing AutoGluon with UniAutoML (with and without LLM-Explainer). }
  \renewcommand{\arraystretch}{1.7}
  \label{hypo}
  \centering
  \resizebox{\columnwidth}{!}{
    \begin{tabular}{llll}
                   
      \cmidrule(r){1-4}
      Hypothesis    & T-value     & P-value & Null Hypothesis \\
      \midrule
      \#1 & 0.404, -8.138  & 0.344, 1.595\(\times 10^{-8}\) & Reject   \\
      \#2 & 0.508, 2.634   & 0.307,  7.407\(\times 10^{-3}\) & Reject   \\
      \#3 & 0.309, 7.207  & 0.379, 9.493\(\times 10^{-8}\) & Reject   \\
      \#4 & 5.608, -4.174  & 5.227\(\times 10^{-6}\), 1.695\(\times 10^{-4}\) & Reject \\
      \bottomrule
    \end{tabular}
  }
\end{table}

\paragraph{Case Study}
To better evaluate the operational differences between UniAutoML and traditional AutoML framework, we conducted a case study. 
Fig.~\ref{case} illustrates fragments of the general process for both UniAutoML and AutoGluon.
%
%
As shown in Fig.~\ref{case} b, the human-centered design of UniAutoML allows users to intervene at multiple points throughout the AutoML pipeline. 
Each stage is accompanied by LLM-generated explanations, enhancing transparency and making the process more accessible to users with varying levels of expertise. 
This case study underscores UniAutoML's emphasis on user engagement, interpretability, and safety in the AutoML process, contrasting with the more opaque, fully automated approach of traditional frameworks like AutoGluon.

\section{Discussion and Conclusion}
\paragraph{Discussion}
UniAutoML represents an advancement in AutoML, addressing limitations of existing AutoML frameworks by unifying both discriminative and generative tasks. 
The integration of LLMs enhances the framework's ability to handle complex, multimodal datasets and enables more intuitive user interactions.
The human-centered design, featuring a conversational interface and real-time explanations, makes ML more accessible to non-experts. 
As AI continues to evolve, methods like UniAutoML that prioritize user engagement, transparency, and versatility will becoming more important in democratizing ML and shaping its future applications.

\paragraph{Limitations and Future Directions}
Despite the promising results of UniAutoML, several important limitations must be addressed. 
Firstly, the financial implications of using state-of-the-art LLMs like GPT-4o cannot be overlooked. 
The substantial costs associated with API access for complex AutoML tasks may prove prohibitive for many users, particularly in resource-constrained environments. 
To address this, future work should explore more cost-effective alternatives, such as efficient local deployment of LLMs, optimization of prompts for conciseness, and the potential use of smaller, task-specific language models fine-tuned for AutoML tasks.
Another limitation is the current implementation's heavy reliance on internet connectivity for both LLM usage and retrieval of pre-trained models.
This dependency restricts the framework's applicability in offline or low-connectivity environments, which are common in many real-world scenarios. 
Future research can also focus on developing strategies for local deployment of LLMs and pre-trained models, creating compact, offline-capable versions of the framework, and implementing caching mechanisms to reduce the need for constant network access.

\paragraph{Conclusion}
UniAutoML automates the entire ML pipeline while providing unprecedented user interactivity and interpretability through CUI and real-time explanations. 
Our comprehensive evaluation, including quantitative experiments and a user study, demonstrates that UniAutoML not only achieves competitive or superior performance compared to traditional AutoML frameworks but also enhances user experience, control, and trust. 
By reducing the cognitive burden on users and making advanced ML methods more accessible, UniAutoML represents a step towards democratizing AI. 
%

\medskip{
\small
\bibliography{cites}

}


\appendix

\section{Prompts}
\label{prompt}
This section presents the prompts used in different modules of UniAutoML to guide the LLM.
The color scheme in these prompts serves several purposes to enhance readability and highlight important information:
\begin{itemize}
    \item \textcolor{blue!70!black}{Blue text} is used for headings and main instructions. 
    
    \item \textcolor{orange}{Orange text} highlights variable or injectable content. It's used for placeholders like \texttt{\{modality\}}, \texttt{\{configs\}}, or \texttt{\{configuration\_data\}}. 
    
    
    \item Black text on the light background is used for the main content.
\end{itemize}

\paragraph{Prompt for the Modality Inference in Model Selection Module}
The following prompt corresponds to the model selection module to accurately identify and categorize the data modalities present in the input dataset.

\begin{tcolorbox}[
enhanced,
colback=azure!10!white,
colframe=azure!50!black,
colbacktitle=azure!40!white,
coltitle=black,
fonttitle=\bfseries\sffamily,
title=Prompt for Modality Inference in Model Selection Module,
arc=8pt,
boxrule=0.8pt,
drop fuzzy shadow
]
\textcolor{blue!70!black}{\textbf{Role:}} You are an AI assistant specializing in analyzing data modalities from multimodal data. \\
\textcolor{blue!70!black}{\textbf{Task:}} Identify the data type of each column within a \texttt{pandas.DataFrame} provided as {\textcolor{orange}{\textbf{dataset}}}. \\
\textcolor{blue!70!black}{\textbf{Output Format:}} Strictly adhere to the following JSON format without additional context:
\texttt{\{"column\_name": "data\_type"\}}\\
\textcolor{blue!70!black}{\textbf{Instructions:}}
\begin{itemize}
\item Determine data types based on column names, data content, and user-provided context, which may include task or dataset context.
\item Ensure all columns are analyzed and included in the output.
\item Identify the label column (typically present as the target variable for prediction or classification).
\end{itemize}

\textcolor{blue!70!black}{\textbf{Modality Examples:}} text, image, audio, video, document, table, semantic\_seg\_img, ner, categorical, numerical, label.

\textcolor{blue!70!black}{\textbf{Reference Examples:}}
\begin{enumerate}
\item Input: instructions: {\textcolor{orange}{\textbf{description\_A}}}, Data: {\textcolor{orange}{\textbf{input\_A}}}
Output: {\textcolor{orange}{\textbf{output\_A}}}
\item Input: instructions: {\textcolor{orange}{\textbf{description\_B}}}, Data: {\textcolor{orange}{\textbf{input\_B}}}
Output: {\textcolor{orange}{\textbf{output\_B}}}
\item Input: instructions: {\textcolor{orange}{\textbf{description\_C}}}, Data: {\textcolor{orange}{\textbf{input\_C}}}
Output: {\textcolor{orange}{\textbf{output\_C}}}
\end{enumerate}

\textcolor{blue!70!black}{\textbf{Your Task:}}
Input: instructions: {\textcolor{orange}{\textbf{description\_D}}}, Data: {\textcolor{orange}{\textbf{input\_D}}}\\
Output:
\end{tcolorbox}

\paragraph{Prompt for the Preprocessing Module}
The following prompt guides the LLM in generating code for data processors in UniAutoML.

\begin{tcolorbox}[
enhanced,
colback=azure!10!white,
colframe=azure!50!black,
colbacktitle=azure!40!white,
coltitle=black,
fonttitle=\bfseries\sffamily,
title=Prompt for Data Preprocessing Module,
arc=8pt,
boxrule=0.8pt,
drop fuzzy shadow
]
\textcolor{blue!70!black}{\textbf{Role:}} You are an AI assistant specializing in writing data processor code in an AutoML task.

\textcolor{blue!70!black}{\textbf{Task:}} Write a function to return the corresponding data processors based on the model's configuration.

\textcolor{blue!70!black}{\textbf{Output Format:}} Strictly adhere to the following dict format:
\texttt{\{"data\_processor\_codes": "codes", "reason": "reason for choosing the data processor"\}}

\textcolor{blue!70!black}{\textbf{Instructions:}}
\begin{itemize}
    \item Follow the provided function structure and naming conventions.
    \item Modify the code based on the dataset modality {\textcolor{orange}{\textbf{\{modality\}}}}, processing only the modalities present.
    \item Load training data into each processor from the global 'dataset' variable.
    \item Include a label data processor for each model.
    \item Write only the code for defining the processor function, not for executing it.
    \item For image processors, set \texttt{train\_transforms = val\_transforms} based on {\textcolor{orange}{\textbf{\{image\_cfg\}}}} and your knowledge.
    \item For semantic segmentation image processors, assign \texttt{img\_transforms} and \texttt{gt\_transforms} based on {\textcolor{orange}{\textbf{\{semantic\_seg\_img\_cfg\}}}} and your knowledge.
\end{itemize}
\end{tcolorbox}

\begin{tcolorbox}[
enhanced,
colback=azure!10!white,
colframe=azure!50!black,
colbacktitle=azure!40!white,
coltitle=black,
fonttitle=\bfseries\sffamily,
arc=8pt,
boxrule=0.8pt,
drop fuzzy shadow
]
\textcolor{blue!70!black}{\textbf{Code Structure:}}\\

\begin{lstlisting}[
frame=single,
framesep=2mm,
backgroundcolor=\color{gray!10},
basicstyle=\footnotesize\ttfamily,
numbers=left,
numberstyle=\tiny\color{gray!50},
]
from autogluon.multimodal.data import process_numerical, process_categorical, process_document, process_image, process_label, process_ner, process_text, process_semantic_seg_img
# Import only the libraries you will use

def processor(modality):
    processed_data = []
    
    # Example for numerical data
    numerical_processor = process_numerical.NumericalProcessor(
        model=numerical_model,
        requires_column_info=False
    )
    numerical_features = {}
    for column, column_type in modality.items():
        if column_type == "numerical":
            numerical_features[column] = dataset[column]
    processed_data.append(
        numerical_processor(
            numerical_features,
            feature_modalities=modality,
            is_training=True
        )
    )
    
    # Add processors for other modalities here
    
    return processed_data
\end{lstlisting}

\textcolor{blue!70!black}{\textbf{Your Task:}} Complete the \texttt{processor} function by adding processors for all modalities present in {\textcolor{orange}{\textbf{\{modality\}}}}.
\end{tcolorbox}
\paragraph{Prompt for Generation of Fusion Model}
The following prompt guides the LLM in generating code for fusion model
\begin{tcolorbox}[
enhanced,
colback=azure!10!white,
colframe=azure!50!black,
colbacktitle=azure!40!white,
coltitle=black,
fonttitle=\bfseries\sffamily,
title=Prompt for Fusion Model Generation,
arc=8pt,
boxrule=0.8pt,
drop fuzzy shadow
]
\textcolor{blue!70!black}{\textbf{Role:}} You are a helpful assistant that writes the Deep learning model code. \\
\textcolor{blue!70!black}{\textbf{Task:}} You task is to only write a fusion model to fuse different base models' features without any explanation. Use \# before every line except the python code.\\
\textcolor{blue!70!black}{\textbf{Instructions:}}
\begin{itemize}
\item Here are some model code for your reference:
Given some base models' config as follow:{\textcolor{orange}{\textbf{\{base\_configs\}}}};
    Give the fusion model config as follow: {\textcolor{orange}{\textbf{\{fusion\_config\}}}}
    You should then respond to me the code with:
   \item Fusion technique should be learnable, MLP is recommended.
   \item The fusion model structure should be defined as fusion
    model and fusion head,which output features and logits,
    respectively.
   \item Base models instance should be defined in Fusion model
    Class.You should not change the value of the output of base
    model instances.
   \item All base models have a uniform variable(
    self.out features dim) to represent the output
    features dimension.
    \item Finding the maximum dimension of all base models'
    output features, and define learnable linear layers to adapt
    all base models' output features to the maximum dimension
    as the input of fusion model. For example, if three models
    have feature dimensions are [512, 768, 64], it will linearly
    map all the features to dimension 768.
    \item Output the logits,features,loss weights of fusion model
    and base models.The return must be in a JSON format:
    {model name:{“logits”:...,“features”:...,“weight”:...}}.
    \item All the network layers and variable
    self.model name,self.loss weight should be defined in
    function init , not in function forward.
    \item Some variables are not present in each model’s config,
    you cannot use a variable that does not exist in the corresponding
    model config.
\end{itemize}

\end{tcolorbox}
\begin{tcolorbox}[
enhanced,
colback=azure!10!white,
colframe=azure!50!black,
colbacktitle=azure!40!white,
coltitle=black,
fonttitle=\bfseries\sffamily,
arc=8pt,
boxrule=0.8pt,
drop fuzzy shadow
]
\textcolor{blue!70!black}{\textbf{Code Structure:}}\\

\begin{lstlisting}[
frame=single,
framesep=2mm,
backgroundcolor=\color{gray!10},
basicstyle=\footnotesize\ttfamily,
numbers=left,
numberstyle=\tiny\color{gray!50},
]
from multimodal.models import CategoricalTransformer
    class CategoricalTransformer(nn.Module):
    def init (self,model config):
    ...
    from multimodal.models import NumericalTransformer
    class NumericalTransformer(nn.Module):
    def init (self,model config):
    ...
    from multimodal.models import TimmAutoModel\
    ForImagePrediction
    class TimmAutoModelFor\
    ImagePrediction(nn.Module):
    def init (self,model config):
    ...
    from multimodal.models import HFAutoModelFor\
    TextPrediction
    class HFAutoModelForTextPrediction(nn.Module):
    def init (self,model config):
    ...
\end{lstlisting}
\end{tcolorbox}
\paragraph{Prompt for the Hyperparameter Description in Model Training Module}

The following prompt guides the LLM in generating descriptions for hyperparameters.

\begin{tcolorbox}[
enhanced,
colback=azure!10!white,
colframe=azure!50!black,
colbacktitle=azure!40!white,
coltitle=black,
fonttitle=\bfseries\sffamily,
title=Prompt for Hyperparameter Description Generation in Model Training Module,
arc=8pt,
boxrule=0.8pt,
drop fuzzy shadow
]
\textcolor{blue!70!black}{\textbf{Role:}} You are an AI assistant specializing in describing machine learning hyperparameters. \\

\textcolor{blue!70!black}{\textbf{Task:}} Add descriptions for the parameters in a machine learning training configuration. \\

\textcolor{blue!70!black}{\textbf{Output Format:}} Strictly adhere to the following JSON format without additional content:
\texttt{\{"hyperparameter\_name": "description"\}}\\

\textcolor{blue!70!black}{\textbf{Instructions:}}
\begin{itemize}
\item Provide clear and concise descriptions for each hyperparameter.
\item Ensure that descriptions do not include specific values from the configuration.
\item Focus on the purpose and impact of each hyperparameter on the model's behavior or performance.
\item Use technical language appropriate for machine learning practitioners.
\end{itemize}

\textcolor{blue!70!black}{\textbf{Given Information:}}
Model configurations: {\textcolor{orange}{\textbf{{configs}}}}\\

\textcolor{blue!70!black}{\textbf{Your Task:}} Generate descriptions for all hyperparameters present in the given model configurations.
\end{tcolorbox}



\paragraph{Prompt for the Hyperparameter Optimization in Model Training Module}
The following prompt guides the LLM in inferring hyperparameters and their search ranges for optimization in the model training module.
\begin{tcolorbox}[
enhanced,
colback=azure!10!white,
colframe=azure!50!black,
colbacktitle=azure!40!white,
coltitle=black,
fonttitle=\bfseries\sffamily,
title=Prompt for Hyperparameter Optimization in Model Training Module,
arc=8pt,
boxrule=0.8pt,
drop fuzzy shadow
]
\textcolor{blue!70!black}{\textbf{Role:}} You are an AI assistant specializing in hyperparameter optimization for machine learning tasks. \\

\textcolor{blue!70!black}{\textbf{Task:}} Infer hyperparameters and their search ranges for optimization.\\

\textcolor{blue!70!black}{\textbf{Output Format:}} Strictly adhere to the following JSON format:
\texttt{{"parameter\_name": "range\_of\_values"}}\\

\textcolor{blue!70!black}{\textbf{Instructions:}}
\begin{itemize}
\item Use the format [option1, option2, option3, ...] to represent a discrete search range.
\item For INT or FLOAT type values, include at least 3 values in the search space.
\item Ensure search ranges include and refer to the original config values.
\item Only output parameters that need optimization.
\item Do not invent parameters not present in the configuration file.
\item If "checkpoint\_identifier" is in the config, only consider the "weight\_loss".
\item If unable to infer search ranges, estimate based on parameter names and general search ranges.
\end{itemize}

\textcolor{blue!70!black}{\textbf{Given Information:}}
Parameter descriptions: {\textcolor{orange}{\textbf{{self\_description}}}}
Configuration data: {\textcolor{orange}{\textbf{{configuration\_data}}}}\\

\textcolor{blue!70!black}{\textbf{Your Task:}} Infer and provide search ranges for all relevant hyperparameters in the given configuration.
\end{tcolorbox}











\paragraph{Prompt for Model Selection Module for UniAutoML}
The following prompt guides the LLM to determine the task type (generative v.s. discriminative) of UniAutoML.
\begin{tcolorbox}[
enhanced,
colback=azure!10!white,
colframe=azure!50!black,
colbacktitle=azure!40!white,
coltitle=black,
fonttitle=\bfseries\sffamily,
title=Prompt for Task Type Selection,
arc=8pt,
boxrule=0.8pt,
drop fuzzy shadow
]
\textcolor{blue!70!black}{\textbf{Role:}} You are an assistant that helps user to switch mode of a model.\\
\textcolor{blue!70!black}{\textbf{Task:}} User will give you its task requiremeng for that AutoML model. Your task is to respond to user input by classifying it.\\
\textcolor{blue!70!black}{\textbf{Instructions:}}
\begin{itemize}
\item If the user wants to conduct discriminative task, respond discriminative
\item If the user wants to conduct task using Diffusion model, respond diffusion
\item If the user wants to conduct task use fine-tuned LLM, respond LLM
\end{itemize}
\end{tcolorbox}
\begin{tcolorbox}[
enhanced,
colback=azure!10!white,
colframe=azure!50!black,
colbacktitle=azure!40!white,
coltitle=black,
fonttitle=\bfseries\sffamily,
title=Prompt for Auto Diffusion,
arc=8pt,
boxrule=0.8pt,
drop fuzzy shadow
]
\textcolor{blue!70!black}{\textbf{Role:}} You are a code engineer that helps user to 
write code to use Diffusion model\\
\textcolor{blue!70!black}{\textbf{Task:}} Your task is to first infer Diffusion model, then fine tune it use LoRA and train dataset which is stored at {\textcolor{orange}{\textbf{\{dataset\_address\}}}}.
\textcolor{blue!70!black}{\textbf{Instructions:}}
\begin{itemize}
\item Here are some code for you to refer to, it is an example of how to infer this model
\end{itemize}
\end{tcolorbox}
\paragraph{Prompt for Fine-Tuning Diffusion Models in the Model Training Module}
The following prompt guides the LLM in fine-tining diffusion models based on dataset provided by the user.
\begin{tcolorbox}[
enhanced,
colback=azure!10!white,
colframe=azure!50!black,
colbacktitle=azure!40!white,
coltitle=black,
fonttitle=\bfseries\sffamily,
arc=8pt,
boxrule=0.8pt,
drop fuzzy shadow
]
\textcolor{blue!70!black}{\textbf{Code Structure:}}\\

\begin{lstlisting}[
frame=single,
framesep=2mm,
backgroundcolor=\color{gray!10},
basicstyle=\footnotesize\ttfamily,
numbers=left,
numberstyle=\tiny\color{gray!50},
]
import torch
import numpy as np
from PIL import Image

from transformers import DPTFeatureExtractor, DPTForDepthEstimation
from diffusers import ControlNetModel, StableDiffusion\
XLControlNetPipeline, AutoencoderKL
from diffusers.utils import load_image


depth_estimator = DPTForDepthEstimation.from_pretrained("Intel/dpt-hybrid-midas").to("cuda")
feature_extractor = DPTFeatureExtractor.from_pretrained("Intel/dpt-hybrid-midas")
controlnet = ControlNetModel.from_pretrained(
    "diffusers/controlnet-depth-sdxl-1.0",
    variant="fp16",
    use_safetensors=True,
    torch_dtype=torch.float16,
)
vae = AutoencoderKL.from_pretrained("madebyollin/sdxl-vae-fp16-fix", torch_dtype=torch.float16)
pipe = StableDiffusionXLControl\
NetPipeline.from_pretrained(
    "stabilityai/stable-diffusion-xl-base-1.0",
    controlnet=controlnet,
    vae=vae,
    variant="fp16",
    use_safetensors=True,
    torch_dtype=torch.float16,
)
pipe.enable_model_cpu_offload()




\end{lstlisting}
\end{tcolorbox}
\begin{tcolorbox}[
enhanced,
colback=azure!10!white,
colframe=azure!50!black,
colbacktitle=azure!40!white,
coltitle=black,
fonttitle=\bfseries\sffamily,
arc=8pt,
boxrule=0.8pt,
drop fuzzy shadow
]

\begin{lstlisting}[
frame=single,
framesep=2mm,
backgroundcolor=\color{gray!10},
basicstyle=\footnotesize\ttfamily,
numbers=left,
numberstyle=\tiny\color{gray!50},
]
def get_depth_map(image):
    image = feature_extractor(images=image, return_tensors="pt").pixel_values.to("cuda")
    with torch.no_grad(), torch.autocast("cuda"):
        depth_map = depth_estimator(image).predicted_depth

    depth_map = torch.nn.functional.interpolate(
        depth_map.unsqueeze(1),
        size=(1024, 1024),
        mode="bicubic",
        align_corners=False,
    )
    depth_min = torch.amin(depth_map, dim=[1, 2, 3], keepdim=True)
    depth_max = torch.amax(depth_map, dim=[1, 2, 3], keepdim=True)
    depth_map = (depth_map - depth_min) / (depth_max - depth_min)
    image = torch.cat([depth_map] * 3, dim=1)

    image = image.permute(0, 2, 3, 1).cpu().numpy()[0]
    image = Image.fromarray((image * 255.0).clip(0, 255).astype(np.uint8))
    return image


prompt = "stormtrooper lecture, photorealistic"
image = load_image("https://huggingface.co/lllyasviel/sd-controlnet-depth/resolve/main/images/stormtrooper.png")
controlnet_conditioning_scale = 0.5  # recommended for good generalization

depth_image = get_depth_map(image)

images = pipe(
    prompt, image=depth_image, num_inference_steps=30, controlnet_conditioning_scale=controlnet_conditioning_scale,
).images
images[0]

images[0].save(f"stormtrooper.png")
\end{lstlisting}
\end{tcolorbox}
\paragraph{Prompt for Preprocessing Dataset with XTuner}
The following prompt guides the LLM in generating data to preprcocess raw data for fine-tuning

\begin{tcolorbox}[
enhanced,
colback=azure!10!white,
colframe=azure!50!black,
colbacktitle=azure!40!white,
coltitle=black,
fonttitle=\bfseries\sffamily,
title=Prompt for Preprocessing Dataset,
arc=8pt,
boxrule=0.8pt,
drop fuzzy shadow
]
\textcolor{blue!70!black}{\textbf{Role:}} You are a code engineer that helps user to preprocess dataset.\\
\textcolor{blue!70!black}{\textbf{Task:}} Preprocess dataset.\\
\textcolor{blue!70!black}{\textbf{Instructions:}}
\begin{itemize}
\item  Map the original dataset to standard form.
\item  Here are example codes for you to refer, it includes codes for both single-turn dialogue pipeline and multi-round dialogue pipeline, judge which to use based on user prompt.
\end{itemize}
\end{tcolorbox}
\begin{tcolorbox}[
enhanced,
colback=azure!10!white,
colframe=azure!50!black,
colbacktitle=azure!40!white,
coltitle=black,
fonttitle=\bfseries\sffamily,
arc=8pt,
boxrule=0.8pt,
drop fuzzy shadow
]

\begin{lstlisting}[
frame=single,
framesep=2mm,
backgroundcolor=\color{gray!10},
basicstyle=\footnotesize\ttfamily,
numbers=left,
numberstyle=\tiny\color{gray!50},
]


# For singe-turn dialogue 
from datasets import load_dataset
ds = load_dataset(path='tatsu-lab/alpaca')
 # Suppose the function is stored in ./map_fn.py
SYSTEM_ALPACA = ('Below is an instruction that describes a task. '
                 'Write a response that appropriately completes the request.\n')
def custom_map_fn(example):
    if example.get('output') == '<nooutput>':
        return {'conversation': []}
    else:
        return {
            'conversation': [{
                'system': SYSTEM_ALPACA,
                'input': f"{example['instruction']}\n{example['input']}",
                'output': example['output']
            }]
        }
#For multipe-turnn dialogue
from datasets import load_dataset

ds = load_dataset(path='timdettmers/openassistant-guanaco')
# Suppose the function is stored in ./map_fn.py
SYSTEM_OASST1 = ''  # oasst1 does not set the system text
def custom_map_fn(example):
    r"""
    Example before preprocessing:
        ...
    """
\end{lstlisting}
\end{tcolorbox}
\begin{tcolorbox}[
enhanced,
colback=azure!10!white,
colframe=azure!50!black,
colbacktitle=azure!40!white,
coltitle=black,
fonttitle=\bfseries\sffamily,
arc=8pt,
boxrule=0.8pt,
drop fuzzy shadow
]

\begin{lstlisting}[
frame=single,
framesep=2mm,
backgroundcolor=\color{gray!10},
basicstyle=\footnotesize\ttfamily,
numbers=left,
numberstyle=\tiny\color{gray!50},
]
 data = []
    for sentence in example['text'].strip().split('###'):
        sentence = sentence.strip()
        if sentence[:6] == 'Human:':
            data.append(sentence[6:].strip())
        elif sentence[:10] == 'Assistant:':
            data.append(sentence[10:].strip())
    if len(data) % 2:
        # The last round of conversation solely consists of input
        # without any output.
        # Discard the input part of the last round, as this part is ignored in
        # the loss calculation.
        data.pop()
    conversation = []
    for i in range(0, len(data), 2):
        system = SYSTEM_OASST1 if i == 0 else ''
        single_turn_conversation = {
            'system': system,
            'input': data[i],
            'output': data[i + 1]}
        conversation.append(single_turn_conversation)
    return {'conversation': conversation}
\end{lstlisting}
\textcolor{blue!70!black} For both 2 situations, store the standard dataset at {\textcolor{orange}{\textbf{\{data\}}}}.
\end{tcolorbox}
\paragraph{Prompt for Modifying Configurations}
The following guides the LLM in modifying configurations.
\begin{tcolorbox}[
enhanced,
colback=azure!10!white,
colframe=azure!50!black,
colbacktitle=azure!40!white,
coltitle=black,
fonttitle=\bfseries\sffamily,
title=Prompt for Modifying Configuations,
arc=8pt,
boxrule=0.8pt,
drop fuzzy shadow
]
\textcolor{blue!70!black}{\textbf{Role:}} You are an assistant that helps to modify configuration\\
\textcolor{blue!70!black}{\textbf{Task:}} Modifying configuration based on user instruction\\
\textcolor{blue!70!black}{\textbf{Instruction:}}
\begin{itemize}
    \item The address user stores LLM is \texttt{\{LLM\_address\}}.
    \item The address user stores cfg is \texttt{\{cfg\_address\}}.
    \item The address user stores dataset is {\textcolor{orange}{\textbf{\{data\}}}}.
    \item The cfg needed to be modified is \texttt{\{cfg\}}.
    \item Here is an example of how to modify cfg.
\end{itemize}
\end{tcolorbox}
\begin{tcolorbox}[
enhanced,
colback=azure!10!white,
colframe=azure!50!black,
colbacktitle=azure!40!white,
coltitle=black,
fonttitle=\bfseries\sffamily,
arc=8pt,
boxrule=0.8pt,
drop fuzzy shadow
]

\textcolor{blue!70!black}{\textbf{Code Structure:}}\\

\begin{lstlisting}[
frame=single,
framesep=2mm,
backgroundcolor=\color{gray!10},
basicstyle=\footnotesize\ttfamily,
numbers=left,
numberstyle=\tiny\color{gray!50},
]

PART 1  Settings                           

# Model
- pretrained_model_name_or_path = 'internlm/internlm2-7b'
+ pretrained_model_name_or_path = './Shanghai_AI_Laboratory/internlm2-chat-7b'

# Data
- data_path = 'burkelibbey/colors'
+ data_path = './colors/train.jsonl'
- prompt_template = PROMPT_TEMPLATE.default
+ prompt_template = PROMPT_TEMPLATE.internlm2_chat

...
PART 3  Dataset & Dataloader
train_dataset = dict(
    type=process_hf_dataset,
-   dataset=dict(type=load_dataset, path=data_path),
+   dataset=dict(type=load_dataset, path='json', data_files=dict(train=data_path)),
    tokenizer=tokenizer,
    max_length=max_length,
    dataset_map_fn=colors_map_fn,
    template_map_fn=dict(
        type=template_map_fn_factory, template=prompt_template),
    remove_unused_columns=True,
    shuffle_before_pack=True,
    pack_to_max_length=pack_to_max_length)
\end{lstlisting}
\end{tcolorbox}

\paragraph{Prompt for Fine-Tuning LLMs in Model Training Module}
The following prompt guides the LLM in fine-tuning LLM provided by the user.
\begin{tcolorbox}[
enhanced,
colback=azure!10!white,
colframe=azure!50!black,
colbacktitle=azure!40!white,
coltitle=black,
fonttitle=\bfseries\sffamily,
title=Prompt for Fine-Tuning LLMs,
arc=8pt,
boxrule=0.8pt,
drop fuzzy shadow
]
\textcolor{blue!70!black}{\textbf{Role:}} You are a code engineering that helps users to fine tune LLM use XTuner.\\
\textcolor{blue!70!black}{\textbf{Task:}} Fine-tuning LLM that user gives and store it.\\
\textcolor{blue!70!black}{\textbf{Instructions:}}
\begin{itemize}
\item Here is an example of how to use XTuner to fine tune LLM.
\item The address user stores cfg is \texttt{\{cfg\_address\}}
\item The address user stores LLM is \texttt{\{LLM\_address\}}.
\item The address user hopes to store fine-tuned LLM is \texttt{\{address\}}.

\item  Here are example codes
\end{itemize}
\end{tcolorbox}
\begin{tcolorbox}[
enhanced,
colback=azure!10!white,
colframe=azure!50!black,
colbacktitle=azure!40!white,
coltitle=black,
fonttitle=\bfseries\sffamily,
arc=8pt,
boxrule=0.8pt,
drop fuzzy shadow
]
\textcolor{blue!70!black}{\textbf{Code Structure:}}\\

\begin{lstlisting}[
frame=single,
framesep=2mm,
backgroundcolor=\color{gray!10},
basicstyle=\footnotesize\ttfamily,
numbers=left,
numberstyle=\tiny\color{gray!50},
]
xtuner train ${CONFIG_NAME_OR_PATH} --deepspeed deepspeed_zero2
xtuner convert pth_to_hf ${CONFIG_NAME_OR_PATH} ${PTH} ${SAVE_PATH}
\end{lstlisting}
\end{tcolorbox}

\section{Dataset Resources}
To facilitate reproducibility and further research, we provide access links to the datasets used in our quantitative experiments on the discriminative tasks. 

\begin{enumerate}
\item \textbf{PetFinder.my-Pawpularity Contest (PPCD)}: A regression task dataset for predicting pet popularity based on image features.
URL: \url{https://www.kaggle.com/competitions/petfinder-pawpularityscore}
\item \textbf{CH-SIMS}: A multimodal sentiment analysis dataset combining text, audio, and visual features.
URL: \url{https://github.com/thuiar/MMSA}
\item \textbf{PetFinder.my-Adoption Prediction (PAPD)}: A multiclass classification dataset for predicting pet adoption speeds.
URL: \url{https://www.kaggle.com/competitions/petfinder-adoptionprediction}
\item \textbf{PARA (Photo Aesthetics Ranking Archive)}: A regression task dataset for image aesthetics assessment.
URL: \url{https://cv-datasets.institutecv.com/#/data-sets}
\item \textbf{Shopee-Price Match Guarantee (SPMG)}: A product matching dataset for e-commerce applications.
URL: \url{https://www.kaggle.com/competitions/shopee-product-matching}
\item \textbf{Multi-Modal Sarcasm Detection (MMSD)}: A binary classification dataset for detecting sarcasm in multimodal content.
URL: \url{https://github.com/headacheboy/data-of-multimodal-sarcasmdetection}
\item \textbf{Titanic - Machine Learning from Disaster (TMLFDD)}: A binary classification dataset for predicting survival on the Titanic.
URL: \url{https://www.kaggle.com/competitions/titanic/data}
\item \textbf{LMSYS}: A multiclass classification dataset for language model evaluation.
URL: \url{https://www.kaggle.com/competitions/lmsys-chatbot-arena/data}
\end{enumerate}

\section{Questionnaire for User Study}
\label{question}

\paragraph{Questionnaire for User Background Survey}
\begin{enumerate}
\item How old are you? single-choice question

\begin{tabularx}{.5\columnwidth}{cc} 
\tikz\draw (0,0) circle (3pt); & <18 \\
\tikz\draw (0,0) circle (3pt); & 18-25 \\
\tikz\draw (0,0) circle (3pt); & 26-30 \\
\tikz\draw (0,0) circle (3pt); & 30-40 \\
\tikz\draw (0,0) circle (3pt); & >40 \\
\end{tabularx}

\item What is your gender? single-choice question

\begin{tabularx}{.5\columnwidth}{cl} 
\tikz\draw (0,0) circle (3pt); & Female \\
\tikz\draw (0,0) circle (3pt); & Male \\

\tikz\draw (0,0) circle (3pt); & Other:\rule{40mm}{0.4pt} \\
\end{tabularx}

\item What is your highest education qualification? single-choice question

\begin{tabularx}{.5\columnwidth}{cl} 
\tikz\draw (0,0) circle (3pt); & High school or lower \\
\tikz\draw (0,0) circle (3pt); & Bachelor \\

\tikz\draw (0,0) circle (3pt); & Master \\
\tikz\draw (0,0) circle (3pt); & PhD \\

\tikz\draw (0,0) circle (3pt); & Other:\rule{40mm}{0.4pt} \\
\end{tabularx}
\item What is your current job? single-choice question

\begin{tabularx}{.5\columnwidth}{cl} 
\tikz\draw (0,0) circle (3pt); & Engineer \\
\tikz\draw (0,0) circle (3pt); & Software development Engineer \\

\tikz\draw (0,0) circle (3pt); & AI Engineer \\
\tikz\draw (0,0) circle (3pt); & Student \\
\tikz\draw (0,0) circle (3pt); & Researchers \\
\tikz\draw (0,0) circle (3pt); & Doctors/ Medical professional \\

\tikz\draw (0,0) circle (3pt); & Other:\rule{40mm}{0.4pt} \\
\end{tabularx}
\item Do you have basic knowledge of large language models? single-choice question

\begin{tabularx}{.1\columnwidth}{cl} 
\tikz\draw (0,0) circle (3pt); & Yes, I have sufficient knowledge about it \\
\tikz\draw (0,0) circle (3pt); & Yes, I have mastered some basic knowledge about it \\

\tikz\draw (0,0) circle (3pt); & No, I know few about it \\

\end{tabularx}
\item Do you have knowledge about Python? single-choice question

\begin{tabularx}{.5\textwidth}{cl} 
\tikz\draw (0,0) circle (3pt); & Yes, I have sufficient knowledge about it \\
\tikz\draw (0,0) circle (3pt); & Yes, I have mastered some basic knowledge about it \\

\tikz\draw (0,0) circle (3pt); &  No, I know few about it\\

\end{tabularx}
\item Do you have knowledge about terminal operation? single-choice question

\begin{tabularx}{.5\textwidth}{cl} 
\tikz\draw (0,0) circle (3pt); & Yes, I have sufficient knowledge about it \\
\tikz\draw (0,0) circle (3pt); & Yes, I have mastered some basic knowledge about it \\

\tikz\draw (0,0) circle (3pt); &  No, I know few about it\\

\end{tabularx}
\item Do you have knowledge about conversational user interface (CUI)? single-choice question

\begin{tabularx}{.5\textwidth}{cl} 
\tikz\draw (0,0) circle (3pt); & Yes, I have sufficient knowledge about it \\
\tikz\draw (0,0) circle (3pt); & Yes, I have mastered some basic knowledge about it \\

\tikz\draw (0,0) circle (3pt); &  No, I know few about it\\

\end{tabularx}
\item Do you have any experience in using machine learning? single-choice question

\begin{tabularx}{.5\textwidth}{cl} 
\tikz\draw (0,0) circle (3pt); & Yes, I have sufficient knowledge about it \\
\tikz\draw (0,0) circle (3pt); & Yes, I have mastered some basic knowledge about it \\

\tikz\draw (0,0) circle (3pt); &  No, I know few about it\\

\end{tabularx}
\item Do you have any experience in using automated machine learning(AutoML) tool? single-choice question

\begin{tabularx}{.5\textwidth}{cl} 
\tikz\draw (0,0) circle (3pt); & Yes, I frequently use it \\
\tikz\draw (0,0) circle (3pt); & Yes, I have used it \\

\tikz\draw (0,0) circle (3pt); &  No, I have never used it\\

\end{tabularx}
\item Do you know how to use AutoGluon? single-choice question

\begin{tabularx}{.5\textwidth}{cl} 
\tikz\draw (0,0) circle (3pt); & Yes \\

\tikz\draw (0,0) circle (3pt); &  No\\

\end{tabularx}
\item Do you agree to attend this experiment? single-choice question

\begin{tabularx}{.5\textwidth}{cl} 
\tikz\draw (0,0) circle (3pt); & Yes \\

\tikz\draw (0,0) circle (3pt); &  No\\

\end{tabularx}
\item What do you expect AutoML tool can do? :  \rule{40mm}{0.4pt}

\end{enumerate}

\paragraph{Questionnaire in User Study}
\label{after}
\begin{enumerate}
\item How long did you spend to finish the experiment? :  \rule{20mm}{0.4pt} (min)
\item How many attempts did you make before finishing the experiment? :  \rule{20mm}{0.4pt} 
\item Do you master how to use the model now?

\begin{tabular}{rccccccl}
& 1 & 2 & 3 & 4 & 5 & \\
\hline
Not at all & \tikz\draw (0,0) circle (3pt); &
\tikz\draw (0,0) circle (3pt); &
\tikz\draw (0,0) circle (3pt); &
\tikz\draw (0,0) circle (3pt); &
\tikz\draw (0,0) circle (3pt); & & Very familiar \\
\hline
\end{tabular}
\item Do you think you will use the model in the future?

\begin{tabular}{rccccccl}
& 1 & 2 & 3 & 4 & 5 & \\
\hline
Impossible & \tikz\draw (0,0) circle (3pt); &
\tikz\draw (0,0) circle (3pt); &
\tikz\draw (0,0) circle (3pt); &
\tikz\draw (0,0) circle (3pt); &
\tikz\draw (0,0) circle (3pt); & & Possible \\
\hline
\end{tabular}
\item Can you understand the results of output from the model?

\begin{tabular}{rccccccl}
& 1 & 2 & 3 & 4 & 5 & \\
\hline
Disagree & \tikz\draw (0,0) circle (3pt); &
\tikz\draw (0,0) circle (3pt); &
\tikz\draw (0,0) circle (3pt); &
\tikz\draw (0,0) circle (3pt); &
\tikz\draw (0,0) circle (3pt); & &Agree \\
\hline
\end{tabular}
\item Do you think the model is complex or simple?

\begin{tabular}{rccccccl}
& 1 & 2 & 3 & 4 & 5 & \\
\hline
Simple & \tikz\draw (0,0) circle (3pt); &
\tikz\draw (0,0) circle (3pt); &
\tikz\draw (0,0) circle (3pt); &
\tikz\draw (0,0) circle (3pt); &
\tikz\draw (0,0) circle (3pt); & & Complex \\
\hline
\end{tabular}
\item Do you think you need any professional to guide you when using the model?

\begin{tabular}{rccccccl}
& 1 & 2 & 3 & 4 & 5 & \\
\hline
No & \tikz\draw (0,0) circle (3pt); &
\tikz\draw (0,0) circle (3pt); &
\tikz\draw (0,0) circle (3pt); &
\tikz\draw (0,0) circle (3pt); &
\tikz\draw (0,0) circle (3pt); & & Very necessary\\
\hline
\end{tabular}
\item Do you feel confident using the model?

\begin{tabular}{rccccccl}
& 1 & 2 & 3 & 4 & 5 & \\
\hline
Not at all & \tikz\draw (0,0) circle (3pt); &
\tikz\draw (0,0) circle (3pt); &
\tikz\draw (0,0) circle (3pt); &
\tikz\draw (0,0) circle (3pt); &
\tikz\draw (0,0) circle (3pt); & & Very confident\\
\hline
\end{tabular}
\item Do you think you need to learn pre-knowledge before being able to use the model?

\begin{tabular}{rccccccl}
& 1 & 2 & 3 & 4 & 5 & \\
\hline
Not at all & \tikz\draw (0,0) circle (3pt); &
\tikz\draw (0,0) circle (3pt); &
\tikz\draw (0,0) circle (3pt); &
\tikz\draw (0,0) circle (3pt); &
\tikz\draw (0,0) circle (3pt); & & Strongly agree\\
\hline
\end{tabular}
\item Do you think people can learn to use the model very quickly?

\begin{tabular}{rccccccl}
& 1 & 2 & 3 & 4 & 5 & \\
\hline
Impossible & \tikz\draw (0,0) circle (3pt); &
\tikz\draw (0,0) circle (3pt); &
\tikz\draw (0,0) circle (3pt); &
\tikz\draw (0,0) circle (3pt); &
\tikz\draw (0,0) circle (3pt); & & Very possible\\
\hline
\end{tabular}
\item Do you think people can learn to use the model very quickly?

\begin{tabular}{rccccccl}
& 1 & 2 & 3 & 4 & 5 & \\
\hline
Impossible & \tikz\draw (0,0) circle (3pt); &
\tikz\draw (0,0) circle (3pt); &
\tikz\draw (0,0) circle (3pt); &
\tikz\draw (0,0) circle (3pt); &
\tikz\draw (0,0) circle (3pt); & & Very possible\\
\hline

\end{tabular}
\item Do you feel you have control over the model?

\begin{tabular}{rccccccl}
& 1 & 2 & 3 & 4 & 5 & \\
\hline
Not at all & \tikz\draw (0,0) circle (3pt); &
\tikz\draw (0,0) circle (3pt); &
\tikz\draw (0,0) circle (3pt); &
\tikz\draw (0,0) circle (3pt); &
\tikz\draw (0,0) circle (3pt); & & Strongly agree\\
\hline

\end{tabular}
\item Physical requirement: How physically challenging was the task?
Please rate on a scale from 1 to 20, where 1 means very low and 20 means very high, \rule{20mm}{0.4pt}
\item Mental requirement: How mentally challenging was the task?
Please rate on a scale from 1 to 20, where 1 means very low and 20 means very high, \rule{20mm}{0.4pt}
\item Temporal requirement: How fast-paced or rushed was the task?
Please rate on a scale from 1 to 20, where 1 means very low and 20 means very high, \rule{20mm}{0.4pt}
\item Performance: How effective were you in completing the task you were assigned?
Please rate on a scale from 1 to 20, where 1 means very low and 20 means very high, \rule{20mm}{0.4pt}
\item Effort: How much exertion was required to achieve your performance level?
Please rate on a scale from 1 to 20, where 1 means very low and 20 means very high, \rule{20mm}{0.4pt}
\item Frustration: How unsure, demoralized, aggravated, tense, and exasperated were you?
Please rate on a scale from 1 to 20, where 1 means very low and 20 means very high, \rule{20mm}{0.4pt}
\item Primary cause of workload? single-choice question\\
\begin{tabular}{cccc}
\tikz\draw (0,0) circle (3pt); & Physical requirement 
&\tikz\draw (0,0) circle (3pt); & Mental requirement \\
\end{tabular}
\item Primary cause of workload? single-choice question\\
\begin{tabular}{cccc}
\tikz\draw (0,0) circle (3pt); & Physical requirement 
&\tikz\draw (0,0) circle (3pt); & Temporal requirement \\
\end{tabular}
\item Primary cause of workload? single-choice question\\
\begin{tabular}{cccc}
\tikz\draw (0,0) circle (3pt); & Physical requirement 
&\tikz\draw (0,0) circle (3pt); & Performance \\
\end{tabular}
\item Primary cause of workload? single-choice question\\
\begin{tabular}{cccc}
\tikz\draw (0,0) circle (3pt); & Physical requirement 
&\tikz\draw (0,0) circle (3pt); & Effort \\
\end{tabular}
\item Primary cause of workload? single-choice question\\
\begin{tabular}{cccc}
\tikz\draw (0,0) circle (3pt); & Physical requirement 
&\tikz\draw (0,0) circle (3pt); & Frustration \\
\end{tabular}
\item Primary cause of workload? single-choice question\\
\begin{tabular}{cccc}
\tikz\draw (0,0) circle (3pt); & Temporal requirement 
&\tikz\draw (0,0) circle (3pt); & Mental requirement \\
\end{tabular}
\item Primary cause of workload? single-choice question\\
\begin{tabular}{cccc}
\tikz\draw (0,0) circle (3pt); &  Performance
&\tikz\draw (0,0) circle (3pt); & Mental requirement \\
\end{tabular}
\item Primary cause of workload? single-choice question\\
\begin{tabular}{cccc}
\tikz\draw (0,0) circle (3pt); &  Effort 
&\tikz\draw (0,0) circle (3pt); & Mental requirement \\
\end{tabular}
\item Primary cause of workload? single-choice question\\
\begin{tabular}{cccc}
\tikz\draw (0,0) circle (3pt); & Frustration 
&\tikz\draw (0,0) circle (3pt); & Mental requirement \\
\end{tabular}
\item Primary cause of workload? single-choice question\\
\begin{tabular}{cccc}
\tikz\draw (0,0) circle (3pt); & Temporal requirement 
&\tikz\draw (0,0) circle (3pt); & Performance \\
\end{tabular}
\item Primary cause of workload? single-choice question\\
\begin{tabular}{cccc}
\tikz\draw (0,0) circle (3pt); & Temporal requirement
&\tikz\draw (0,0) circle (3pt); & Effort \\
\end{tabular}
\item Primary cause of workload? single-choice question\\
\begin{tabular}{cccc}
\tikz\draw (0,0) circle (3pt); &  Temporal requirement
&\tikz\draw (0,0) circle (3pt); &Frustration\\
\end{tabular}
\item Primary cause of workload? single-choice question\\
\begin{tabular}{cccc}
\tikz\draw (0,0) circle (3pt); & Performance
&\tikz\draw (0,0) circle (3pt); & Effort \\
\end{tabular}
\item Primary cause of workload? single-choice question\\
\begin{tabular}{cccc}
\tikz\draw (0,0) circle (3pt); & Physical requirement 
&\tikz\draw (0,0) circle (3pt); & Frustration \\
\end{tabular}
\item Primary cause of workload? single-choice question\\
\begin{tabular}{cccc}
\tikz\draw (0,0) circle (3pt); & Effort
&\tikz\draw (0,0) circle (3pt); & Frustration\\
\end{tabular}

\end{enumerate}

\section{Quantitative Results for Single-Modal Datasets on Discriminative Tasks}
While UniAutoML demonstrates proficiency in handling multi-modal datasets, we also evaluated its performance on single-modality datasets to assess its versatility for discriminative tasks. 
We conducted experiments using the AutoML benchmark \cite{gijsbers2019open} from OpenML \footnote{\url{https://www.openml.org/}}, which encompasses a diverse range of tasks including multiclass classification, regression, and binary classification.
To ensure a comprehensive evaluation, we employed task-specific metrics: Root Mean Squared Error (RMSE) for regression tasks, Area Under the Receiver Operating Characteristic Curve (AUC) for binary classification, and Logarithmic Loss (LogLoss) for multiclass classification. 
We compared UniAutoML against AutoGluon \cite{erickson2020autogluon}, Auto-sklearn \cite{feurer-arxiv20a}, GAMA \cite{gijsbers2021gama}, H2O AutoML \cite{ledell2020h2o}, TPOT \cite{olson2016tpot}, LightAutoML \cite{vakhrushev2021lightautoml}, and FLAML \cite{wang2021flaml}.
Consistent with our previous experimental setup, we configured UniAutoML to prioritize high training quality to maximize accuracy. 
Our evaluation encompassed 24 distinct datasets, with results presented in Tables \ref{long1} and \ref{long2}.
UniAutoML outperformed competing models in 12 out of 24 experiments, including datasets such as Australian and Car. 
This performance is particularly noteworthy for binary classification tasks, where UniAutoML consistently demonstrated superior results. 
For the remaining datasets, UniAutoML's performance remained competitive, with only marginal differences compared to the best-performing models.
These findings underscore UniAutoML's robust performance across various single-modality datasets, complementing its capabilities in multi-modal scenarios. 

\begin{table*}[ht]
\centering
\caption{Comparative Performance for Single Modality Datasets Analysis - Part 1 }
\label{long1}
\resizebox{\textwidth}{!}{
\begin{tabular}{lccccccc}
\toprule
\textbf{Task Name} & \textbf{Task Type} & \textbf{Task Metric} & \textbf{AutoGluon} & \textbf{Auto-sklearn} & \textbf{Auto-sklearn 2} & \textbf{FLAML} & \textbf{GAMA} \\
\midrule

australi... & binary & accuracy & $0.940$ {\tiny $\pm$ 0.020} & $0.932$ {\tiny $\pm$ 0.019} & $0.940$ {\tiny $\pm$ 0.020} & $0.939$ {\tiny $\pm$ 0.025} & $0.940$ {\tiny $\pm$ 0.019} \\
wilt & binary & accuracy & $0.994$ {\tiny $\pm$ 0.009} & $0.994$ {\tiny $\pm$ 0.010} & $0.995$ {\tiny $\pm$ 0.008} & $0.988$ {\tiny $\pm$ 0.013} & $0.996$ {\tiny $\pm$ 0.004} \\
numerai2... & binary & accuracy & $0.524$ {\tiny $\pm$ 0.005} & $0.530$ {\tiny $\pm$ 0.005} & $0.531$ {\tiny $\pm$ 0.004} & $0.528$ {\tiny $\pm$ 0.005} & $0.532$ {\tiny $\pm$ 0.004} \\
credit-g & binary & accuracy & $0.791$ {\tiny $\pm$ 0.039} & $0.783$ {\tiny $\pm$ 0.042} & $0.795$ {\tiny $\pm$ 0.038} & $0.784$ {\tiny $\pm$ 0.039} & $0.791$ {\tiny $\pm$ 0.030} \\
apsfailu... & binary & accuracy & $0.992$ {\tiny $\pm$ 0.002} & $0.992$ {\tiny $\pm$ 0.002} & $0.992$ {\tiny $\pm$ 0.003} & $0.992$ {\tiny $\pm$ 0.003} & $0.992$ {\tiny $\pm$ 0.002} \\
ozone-le... & binary & accuracy & $0.934$ {\tiny $\pm$ 0.017} & $0.920$ {\tiny $\pm$ 0.024} & $0.933$ {\tiny $\pm$ 0.022} & $0.925$ {\tiny $\pm$ 0.021} & $0.926$ {\tiny $\pm$ 0.032} \\
ada & binary & accuracy & $0.920$ {\tiny $\pm$ 0.018} & $0.917$ {\tiny $\pm$ 0.017} & $0.920$ {\tiny $\pm$ 0.018} & \textbf{\boldmath$0.924$ {\tiny $\pm$ 0.018}} & $0.921$ {\tiny $\pm$ 0.018} \\
blood-tr... & binary & accuracy & $0.755$ {\tiny $\pm$ 0.044} & $0.745$ {\tiny $\pm$ 0.052} & $0.755$ {\tiny $\pm$ 0.040} & $0.731$ {\tiny $\pm$ 0.066} & $0.757$ {\tiny $\pm$ 0.049} \\
qsar-bio... & binary & accuracy & $0.941$ {\tiny $\pm$ 0.035} & $0.929$ {\tiny $\pm$ 0.036} & $0.937$ {\tiny $\pm$ 0.027} & $0.928$ {\tiny $\pm$ 0.033} & $0.937$ {\tiny $\pm$ 0.032} \\
pc4 & binary & accuracy & $0.951$ {\tiny $\pm$ 0.018} & $0.941$ {\tiny $\pm$ 0.020} & $0.949$ {\tiny $\pm$ 0.017} & $0.949$ {\tiny $\pm$ 0.019} & $0.951$ {\tiny $\pm$ 0.019} \\
kr-vs-kp & binary & accuracy & \textbf{\boldmath$1.000$ {\tiny $\pm$ 0.000}} & \textbf{\boldmath$1.000$ {\tiny $\pm$ 0.000}} & \textbf{\boldmath$1.000$ {\tiny $\pm$ 0.000}} & \textbf{\boldmath$1.000$ {\tiny $\pm$ 0.000}} & \textbf{\boldmath$1.000$ {\tiny $\pm$ 0.000}} \\
quake & regression & rmse & $0.19$ {\tiny $\pm$ 0.0093} & $0.19$ {\tiny $\pm$ 0.0089} & - & $0.19$ {\tiny $\pm$ 0.0091} & $0.19$ {\tiny $\pm$ 0.0092} \\
sensory & regression & rmse & \textbf{\boldmath$0.67$ {\tiny $\pm$ 0.061}} & $0.69$ {\tiny $\pm$ 0.051} & - & $0.69$ {\tiny $\pm$ 0.054} & $0.68$ {\tiny $\pm$ 0.055} \\
space ga & regression & rmse & \textbf{\boldmath$0.094$ {\tiny $\pm$ 0.013}} & $0.1$ {\tiny $\pm$ 0.025} & - & $0.1$ {\tiny $\pm$ 0.015} & $0.096$ {\tiny $\pm$ 0.019} \\
topo21 & regression & rmse & $0.028$ {\tiny $\pm$ 0.0049} & $0.028$ {\tiny $\pm$ 0.0049} & - & $0.028$ {\tiny $\pm$ 0.0048} & $0.028$ {\tiny $\pm$ 0.0048} \\
abalone & regression & rmse & $2.1$ {\tiny $\pm$ 0.12} & $2.1$ {\tiny $\pm$ 0.11} & - & $2.1$ {\tiny $\pm$ 0.12} & \textbf{\boldmath$2.1$ {\tiny $\pm$ 0.1}} \\
pol & regression & rmse & $2.6$ {\tiny $\pm$ 0.29} & $3.3$ {\tiny $\pm$ 0.35} & - & $3.6$ {\tiny $\pm$ 0.37} & $3.7$ {\tiny $\pm$ 0.3} \\
elevators & regression & rmse & \textbf{\boldmath$0.0018$ {\tiny $\pm$ 5.2e-05}} & $0.0019$ {\tiny $\pm$ 7.3e-05} & - & $0.002$ {\tiny $\pm$ 6.5e-05} & $0.0019$ {\tiny $\pm$ 6.5e-05} \\
eucalypt... & multiclass & logloss & $0.690$ {\tiny $\pm$ 0.053} & $0.716$ {\tiny $\pm$ 0.047} & $0.704$ {\tiny $\pm$ 0.061} & $0.779$ {\tiny $\pm$ 0.121} & $0.700$ {\tiny $\pm$ 0.057} \\
yeast & multiclass & logloss & $1.015$ {\tiny $\pm$ 0.087} & $1.043$ {\tiny $\pm$ 0.080} & $1.015$ {\tiny $\pm$ 0.084} & $1.011$ {\tiny $\pm$ 0.083} & $1.019$ {\tiny $\pm$ 0.081} \\
car & multiclass & logloss & $0.004$ {\tiny $\pm$ 0.011} & $0.004$ {\tiny $\pm$ 0.008} & $0.002$ {\tiny $\pm$ 0.004} & $0.003$ {\tiny $\pm$ 0.005} & $0.012$ {\tiny $\pm$ 0.008} \\
dna & multiclass & logloss & \textbf{\boldmath$0.106$ {\tiny $\pm$ 0.027}} & $0.116$ {\tiny $\pm$ 0.032} & $0.111$ {\tiny $\pm$ 0.025} & $0.106$ {\tiny $\pm$ 0.029} & $0.106$ {\tiny $\pm$ 0.028} \\
helena & multiclass & logloss & \textbf{\boldmath$2.470$ {\tiny $\pm$ 0.016}} & $2.526$ {\tiny $\pm$ 0.018} & $2.485$ {\tiny $\pm$ 0.031} & $2.564$ {\tiny $\pm$ 0.019} & $2.731$ (nan) \\
okcupid-... & multiclass & logloss & $0.559$ {\tiny $\pm$ 0.009} & $0.567$ {\tiny $\pm$ 0.007} & $0.563$ {\tiny $\pm$ 0.008} & $0.562$ {\tiny $\pm$ 0.008} & $0.568$ {\tiny $\pm$ 0.007} \\

\bottomrule
\end{tabular}
}
\end{table*}

\begin{table*}[ht]
\centering
\caption{Comparative Performance for Single Modality Datasets Analysis - Part 2}
\label{long2}
\resizebox{\textwidth}{!}{
\begin{tabular}{lccccccc}
\toprule
\textbf{Task Name} & \textbf{Task Type} & \textbf{Task Metric} & \textbf{H2O AUTOML} & \textbf{LIGHT AUTOML} & \textbf{MLJAR} & \textbf{TPOT} & \textbf{UniAutoML} \\
\midrule

australi... & binary & accuracy & $0.932$ {\tiny $\pm$ 0.021} & $0.945$ {\tiny $\pm$ 0.020} & $0.939$ {\tiny $\pm$ 0.023} & $0.937$ {\tiny $\pm$ 0.028} & \textbf{\boldmath$0.963$ {\tiny $\pm$ 0.022}} \\
wilt & binary & accuracy & $0.994$ {\tiny $\pm$ 0.008} & $0.995$ {\tiny $\pm$ 0.006} & $0.996$ {\tiny $\pm$ 0.004} & $0.994$ {\tiny $\pm$ 0.010} & \textbf{\boldmath$0.997$ {\tiny $\pm$ 0.003}} \\
numerai2... & binary & accuracy & $0.533$ {\tiny $\pm$ 0.006} & $0.529$ {\tiny $\pm$ 0.006} & $0.532$ {\tiny $\pm$ 0.004} & $0.526$ {\tiny $\pm$ 0.007} & \textbf{\boldmath$0.536$ {\tiny $\pm$ 0.008}} \\
credit-g & binary & accuracy & $0.781$ {\tiny $\pm$ 0.044} & $0.790$ {\tiny $\pm$ 0.036} & - & $0.789$ {\tiny $\pm$ 0.035} & \textbf{\boldmath$0.822$ {\tiny $\pm$ 0.031}} \\
apsfailu... & binary & accuracy & $0.991$ {\tiny $\pm$ 0.001} & $0.992$ {\tiny $\pm$ \text{nan}} & \textbf{\boldmath$0.994$ {\tiny $\pm$ 0.002}} & $0.991$ {\tiny $\pm$ 0.004} & $0.992$ {\tiny $\pm$ 0.014} \\
ozone-le... & binary & accuracy & $0.929$ {\tiny $\pm$ 0.018} & $0.931$ {\tiny $\pm$ 0.017} & $0.913$ {\tiny $\pm$ 0.020} & $0.919$ {\tiny $\pm$ 0.025} & \textbf{\boldmath$0.949$ {\tiny $\pm$ 0.014}} \\
ada & binary & accuracy & $0.919$ {\tiny $\pm$ 0.016} & $0.923$ {\tiny $\pm$ 0.017} & $0.920$ {\tiny $\pm$ 0.019} & $0.918$ {\tiny $\pm$ 0.019} & $0.921$ {\tiny $\pm$ 0.022} \\
blood-tr... & binary & accuracy & $0.759$ {\tiny $\pm$ 0.030} & $0.748$ {\tiny $\pm$ 0.054} & - & $0.755$ {\tiny $\pm$ 0.042} & \textbf{\boldmath$0.791$ {\tiny $\pm$ 0.046}} \\
qsar-bio... & binary & accuracy & $0.936$ {\tiny $\pm$ 0.038} & $0.934$ {\tiny $\pm$ 0.032} & $0.925$ {\tiny $\pm$ \text{nan}} & $0.930$ {\tiny $\pm$ 0.030} & \textbf{\boldmath$0.948$ {\tiny $\pm$ 0.025}} \\
pc4 & binary & accuracy & $0.947$ {\tiny $\pm$ 0.021} & $0.951$ {\tiny $\pm$ 0.015} & $0.952$ {\tiny $\pm$ 0.018} & $0.944$ {\tiny $\pm$ 0.022} & \textbf{\boldmath$0.959$ {\tiny $\pm$ 0.016}} \\
kr-vs-kp & binary & accuracy & \textbf{\boldmath$1.000$ {\tiny $\pm$ 0.000}} & \textbf{\boldmath$1.000$ {\tiny $\pm$ 0.000}} & \textbf{\boldmath$1.000$ {\tiny $\pm$ 0.000}} & \textbf{\boldmath$1.000$ {\tiny $\pm$ 0.000}} & \textbf{\boldmath$1.000$ {\tiny $\pm$ 0.000}} \\
quake & regression & rmse & $0.191$ {\tiny $\pm$ 0.0095} & $0.192$ {\tiny $\pm$ 0.0100} & $0.190$ {\tiny $\pm$ 0.0092} & $0.192$ {\tiny $\pm$ 0.0095} & \textbf{\boldmath$0.181$ {\tiny $\pm$ 0.0107}} \\
sensory & regression & rmse & $0.71$ {\tiny $\pm$ 0.061} & $0.70$ {\tiny $\pm$ 0.062} & $0.68$ {\tiny $\pm$ 0.044} & $0.69$ {\tiny $\pm$ 0.055} & $0.71$ {\tiny $\pm$ 0.064} \\
space ga & regression & rmse & $0.098$ {\tiny $\pm$ 0.013} & $0.101$ {\tiny $\pm$ 0.018} & $0.100$ {\tiny $\pm$ 0.019} & $0.100$ {\tiny $\pm$ 0.019} & $0.101$ {\tiny $\pm$ 0.012} \\
topo21 & regression & rmse & $0.027$ {\tiny $\pm$ 0.0050} & $0.028$ {\tiny $\pm$ 0.0050} & \textbf{\boldmath$0.027$ {\tiny $\pm$ 0.0049}} & $0.028$ {\tiny $\pm$ 0.0049} & $0.028$ {\tiny $\pm$ 0.0043} \\
abalone & regression & rmse & $2.2$ {\tiny $\pm$ 0.12} & $2.2$ {\tiny $\pm$ 0.13} & $2.2$ {\tiny $\pm$ 0.13} & $2.2$ {\tiny $\pm$ 0.12} & $2.2$ {\tiny $\pm$ 0.11} \\
pol & regression & rmse & $3.5$ {\tiny $\pm$ 0.29} & $3.8$ {\tiny $\pm$ 0.34} & $2.3$ {\tiny $\pm$ 0.24} & $3.6$ {\tiny $\pm$ 0.39} & \textbf{\boldmath$2.3$ {\tiny $\pm$ 0.17}} \\
elevators & regression & rmse & $0.0021$ {\tiny $\pm$ 0.00014} & $0.0021$ {\tiny $\pm$ 5.6e-05} & $0.0020$ {\tiny $\pm$ 5.7e-05} & $0.0020$ {\tiny $\pm$ 6.3e-05} & $0.0019$ {\tiny $\pm$ 6.3e-05} \\
eucalypt... & multiclass & logloss & $0.704$ {\tiny $\pm$ 0.086} & $0.694$ {\tiny $\pm$ 0.057} & \textbf{\boldmath$0.653$ {\tiny $\pm$ 0.053}} & $0.749$ {\tiny $\pm$ 0.128} & $0.701$ {\tiny $\pm$ 0.056} \\
yeast & multiclass & logloss & $1.042$ {\tiny $\pm$ 0.092} & $1.041$ {\tiny $\pm$ 0.095} & \textbf{\boldmath$1.006$ {\tiny $\pm$ 0.086}} & $1.032$ {\tiny $\pm$ 0.082} & $1.024$ {\tiny $\pm$ 0.081} \\
car & multiclass & logloss & $0.002$ {\tiny $\pm$ 0.002} & $0.001$ {\tiny $\pm$ 0.003} & \textbf{\boldmath$0.003$ {\tiny $\pm$ 0.004}} & $1.452$ {\tiny $\pm$ 3.002} & \textbf{\boldmath$0.002$ {\tiny $\pm$ 0.002}} \\
dna & multiclass & logloss & $0.108$ {\tiny $\pm$ 0.031} & $0.110$ {\tiny $\pm$ 0.027} & $0.108$ {\tiny $\pm$ 0.026} & $0.110$ {\tiny $\pm$ 0.027} & $0.108$ {\tiny $\pm$ 0.026} \\
helena & multiclass & logloss & $2.791$ {\tiny $\pm$ 0.019} & $2.502$ {\tiny $\pm$ 0.013} & $2.573$ {\tiny $\pm$ 0.022} & $2.923$ {\tiny $\pm$ 0.038} & $2.542$ {\tiny $\pm$ 0.019} \\
okcupid-... & multiclass & logloss & $0.568$ {\tiny $\pm$ 0.009} & \textbf{\boldmath$0.559$ {\tiny $\pm$ 0.008}} & $0.565$ {\tiny $\pm$ 0.007} & $0.567$ {\tiny $\pm$ 0.010} & $0.564$ {\tiny $\pm$ 0.008} \\

\bottomrule
\end{tabular}
}
\end{table*}

\section{User Study Details}

\subsection{IRB approved user study}
\label{irb}
This user study was conducted with full approval from the Institutional Review Board (IRB). All methodologies, protocols, and procedures involving human participants were reviewed to ensure compliance with ethical standards in research.

\subsection{Participant Recruitment and Demographics}
\label{bg}
We recruited 25 participants with diverse backgrounds to represent potential users of AutoML toolkits. 
The sample included individuals both with and without machine learning experience. Our participant pool consisted of 15 Computer Science students, 5 teachers from non-computer science fields, 2 AI researchers, and 3 communication engineering teachers.
Participant backgrounds were assessed across three key areas: LLM familiarity, machine learning experience, and AutoML experience. In terms of LLM familiarity, 8 users (32\%) reported being very familiar, 14 users (56\%) had used LLMs, and 3 users (12\%) had never used LLMs. Regarding machine learning experience, 5 users (20\%) were very familiar, 5 users (20\%) had written ML code, and 15 users (60\%) had no ML experience. As for AutoML experience, 2 users (8\%) were very familiar, 1 user (4\%) had previous AutoML experience, and 22 users (88\%) had no AutoML experience.
This diverse participant pool enhances the credibility of our study by including users from various domains and with varying levels of AutoML expertise.

\subsection{Variable Definition and Quantitative Calculation of User Feedback}
\paragraph{Variable Definition}
To facilitate quantitative analysis, we defined variables based on user responses. For the $i^{th}$ question in Appendix \ref{after}, the user response is denoted as $SUS_i$.
\paragraph{Quantitative Calculation}
Our questionnaire in Appendix \ref{after} comprises both numerical (questions 1-18) and categorical (questions 19-33) responses. We employed two distinct methods to analyze these different response types.
\paragraph{Calculation of Numerical Responses}
For numerical responses, we adapted the System Usability Scale (SUS) questionnaire \cite{brooke1996sus}. 
This standardized tool uses a 5-point Likert scale for 10 questions (questions 3-12 in our study), ranging from "strongly disagree" to "strongly agree". 
To enhance clarity, we explicitly defined these extremes in each question. 
The usability score is calculated using the following equation, where higher scores indicate better usability:
$Usability = 2.5 * (20 + \sum_{n=1}^{5} (SUS_{2n+1}) - (SUS_{2n}))$

\paragraph{Calculation of Categorical Responses}
For categorical responses, we employed the NASA Task Load Index (NASA-TLX) \cite{hart1988development}. 
This method assesses workload across six dimensions, namely physical demand, mental demand, temporal demand, performance, effort, and frustration. 
Participants first rate each dimension (questions 13-18) and then perform pairwise comparisons to determine relative importance (questions 19-33).
The workload score is calculated as:
$Workload = \sum_{i=13}^{18} w_i * 5 * s_i$
where $w_i$ is the weight of each dimension (occurrences in pairwise comparisons divided by 15), and $s_i$ is the rating for that dimension.

\paragraph{Limitations and Potential Confounds}
While we have strived to quantify all results, we acknowledge that certain independent variables may influence the study outcomes. 
Our participant pool includes individuals of varying ages, educational levels, and professional backgrounds. These demographic factors may indirectly affect performance in the user study and were considered during data analysis and interpretation.

\begin{figure}[htbp]
\centering
\includegraphics[width=\linewidth]{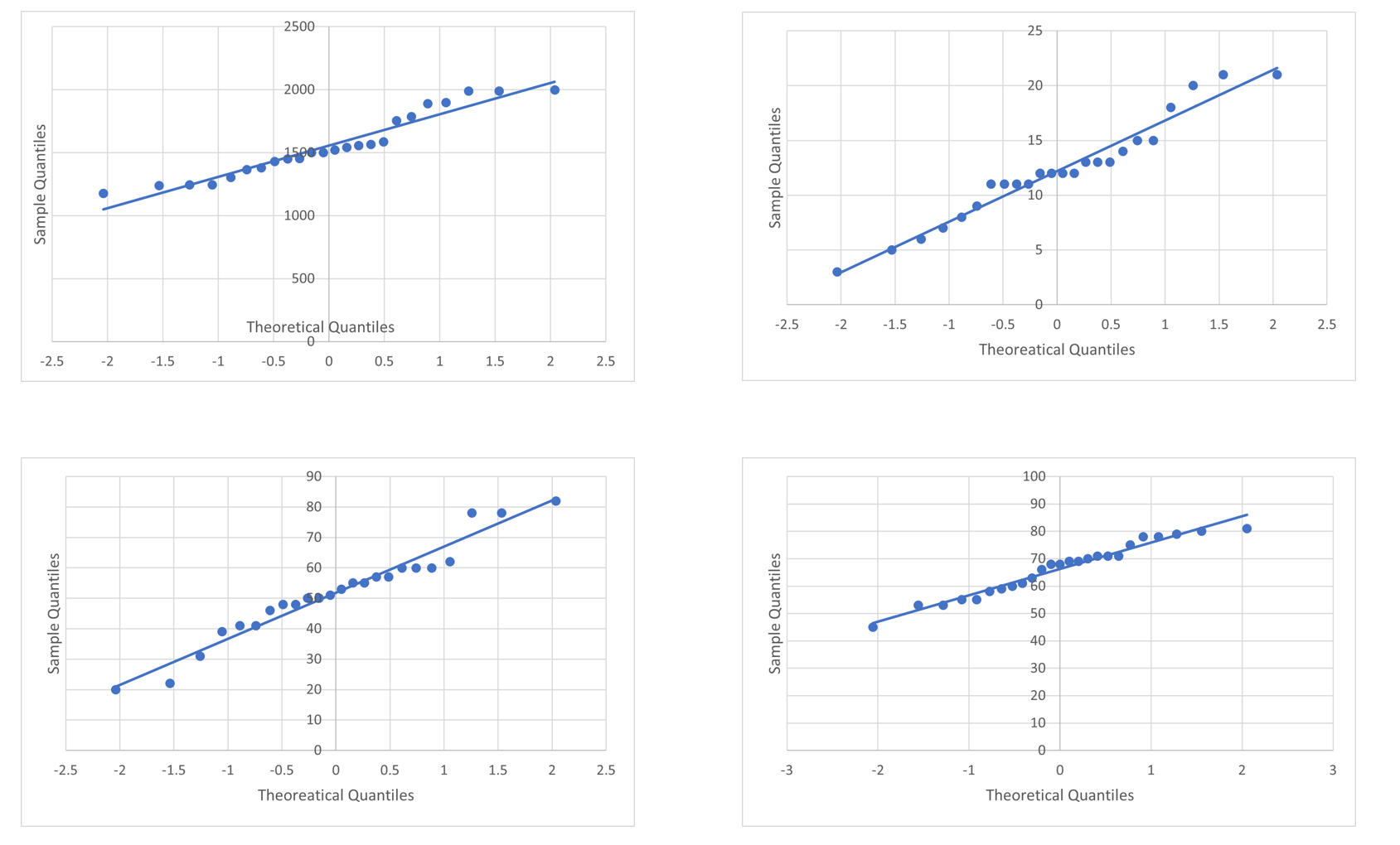}
\caption{QQ plots for all 4 variables, in the order of accomplishing time, attempt times, usability score, workload.}
\label{fig3}
\end{figure}

\subsubsection{User Interviews}
\label{int}
To complement our quantitative analysis and gain deeper insights into user experiences, we conducted interviews with two randomly selected participants. 
These interviews aimed to capture more subjective and personalized feedback.

\paragraph{User 1: Embedded Systems Student}
The first interviewee was a student majoring in embedded processors, who recently participated in a competition requiring real-time facial detection using a provided board. 
As a novice in AI, they found both AutoML frameworks beneficial for their task. However, they noted a significant advantage in UniAutoML's CUI:
``\textit{Using AutoGluon required me to learn some Python coding, which was challenging given the time constraints. UniAutoML's CUI was more suitable for my needs. I could simply type instructions and receive intermediate results with interpretations, without requiring extensive pre-knowledge. This approach was more accessible and efficient for my project.}''

\paragraph{User 2: AI Researcher}
The second interviewee was an experienced AI researcher working at an institution. They provided a perspective from a professional standpoint:
``\textit{I was initially skeptical when I first encountered AutoML frameworks years ago. As a professional AI researcher, I acknowledge their utility for certain tasks, potentially reducing effort in some areas. Personally, the presence of a CUI or interpretations doesn't significantly impact my workflow. However, I recognize the potential benefits for non-professionals or beginners in the field. The concept of making AI more accessible through these interfaces is intriguing and could have significant implications for the broader adoption of AI technologies.}"

These interviews highlight the diverse perspectives on UniAutoML's features, particularly its CUI and LLM-Explainer components. While novice users find these features important for accessibility and ease of use, experienced professionals appreciate their potential for democratizing AI technologies. 
This feedback underscores the importance of designing AutoML systems that cater to users with varying levels of expertise and backgrounds.

\subsection{Process for Analyzing User Study Results}
Our analysis of the user study results followed a structured approach, involving data visualization, normality testing, and hypothesis testing. 
\paragraph{Response Data Visualization}
We began by aggregating responses from all 25 participants for the questions in Appendix \ref{question}. 
The data was visualized using box plots, as shown in Fig.~\ref{fig2}. 
These plots provide a clear representation of the data distribution, displaying the first quartile (Q1), median (Q2), third quartile (Q3), and maximum values for each of the three preset conditions.

\paragraph{Normality Testing}
To validate the statistical assumptions underlying our analysis, we conducted normality tests using Q-Q plots, presented in Fig.~\ref{fig3}. 
These plots compare the distribution of our observed data (y-axis) against the theoretical quantiles of a normal distribution (x-axis), which is important for determining the appropriateness of subsequent parametric statistical tests.

\paragraph{Hypothesis Testing}
To evaluate the statistical significance of observed performance differences between AutoGluon and UniAutoML, we employed hypothesis testing. We calculated the discrepancies ($d_i$) by subtracting UniAutoML measurements from corresponding AutoGluon measurements.
Under the null hypothesis ($H_0$) that AutoGluon and UniAutoML have equivalent effects, these differences should follow a distribution centered around zero ($\mu_d = 0$). 
Our alternative hypothesis ($H_1$) posited that there is a significant difference between the two systems.
For most hypotheses, we tested $H_0: \mu_d = 0$ against $H_1: \mu_d > 0$. 
However, for hypotheses \#3 , we tested against $H_1: \mu_d < 0$, reflecting the specific directionality of these comparisons.
We used a one-sample t-test with the following test statistic:
$t = \frac{\bar{d}}{\frac{s}{\sqrt{n}}}$,
where $\bar{d}$ is the sample mean of the differences, $s$ is the sample standard deviation, and $n$ is the sample size (25 in our case).
The resulting t-value was then compared against the critical value from a t-distribution with 24 degrees of freedom, at the chosen significance level $\alpha$.

\section{Related Works}

\subsection{Automated Machine Learning (AutoML)}
Automated Machine Learning (AutoML) is designed to streamline the complex and repetitive processes associated with traditional ML, enhancing efficiency and productivity. 
AutoML methods typically encompass six crucial stages: (1) data pre-processing, (2) feature selection, (3) model selection, (4) hyperparameter tuning, (5) post-processing, and (6) in-depth assessment and analysis of results.
The core compomet driving AutoML's efficiency include Automated Data Cleaning (Auto Clean) \cite{mahdavi2019towards}, Hyperparameter Optimization (HPO) \cite{feurer2019hyperparameter,yang2020hyperparameter}, Automated Feature Engineering (Auto FE) \cite{khurana2016cognito,zdravevski2017improving}, Meta-Learning, and Neural Architecture Search (NAS) \cite{liu2021survey,ren2021comprehensive}.
Existing AutoML methods like AutoGluon offer automated pipelines that handle model selection, data processing, and hyperparameter optimization for various data types, including tabular, multi-modal, and time-series datasets. 
Other methods such as AutoKeras \cite{JMLR:v24:20-1355} and auto-sklearn employ different model selection methods, with AutoKeras and auto-sklearn utilizing Bayesian Optimization \cite{pelikan2005bayesian}, while AutoGluon leverages stacking ensembles for improved performance.
Recent research has explored the integration of LLMs in AutoML, both for configuring AutoML models \cite{thornton2013auto,zimmer2021auto} and as components within AutoML systems \cite{yu2023gpt,hollmann2024large}. 
%

\subsection{Large Language Model (LLM)} 
LLMs\cite{mann2020language,chowdhery2023palm,chung2024scaling,touvron2023llama,wei2022emergent}, characterized by their immense number of parameters, have demonstrated intelligence and learning capabilities. 
The breakthrough in this domain was facilitated by the introduction of the self-attention mechanism \cite{vaswani2017attention}. 
Various architectures, including the Transformer, have been employed to train these LLMs, with GPT (Generative Pre-trained Transformer)\cite{openai2022chatgpt} being a prominent example.

GPT models excel in natural language processing tasks and possess vast stores of knowledge, making them suitable for generating AutoML code based on both system and user prompts. 
However, the utilization of such powerful models comes with potential risks, such as the generation of harmful or biased information. 
To mitigate these risks, our proposed UniAutoML incorporates a safety guard-line mechanism that scrutinizes user instructions and filters LLM outputs, effectively eliminating harmful or irrelevant information. 
This addition ensures the responsible and safe application of LLMs in creating a robust AutoML model.
\begin{figure}[htbp]
    \centering
    \includegraphics[width=1\columnwidth]{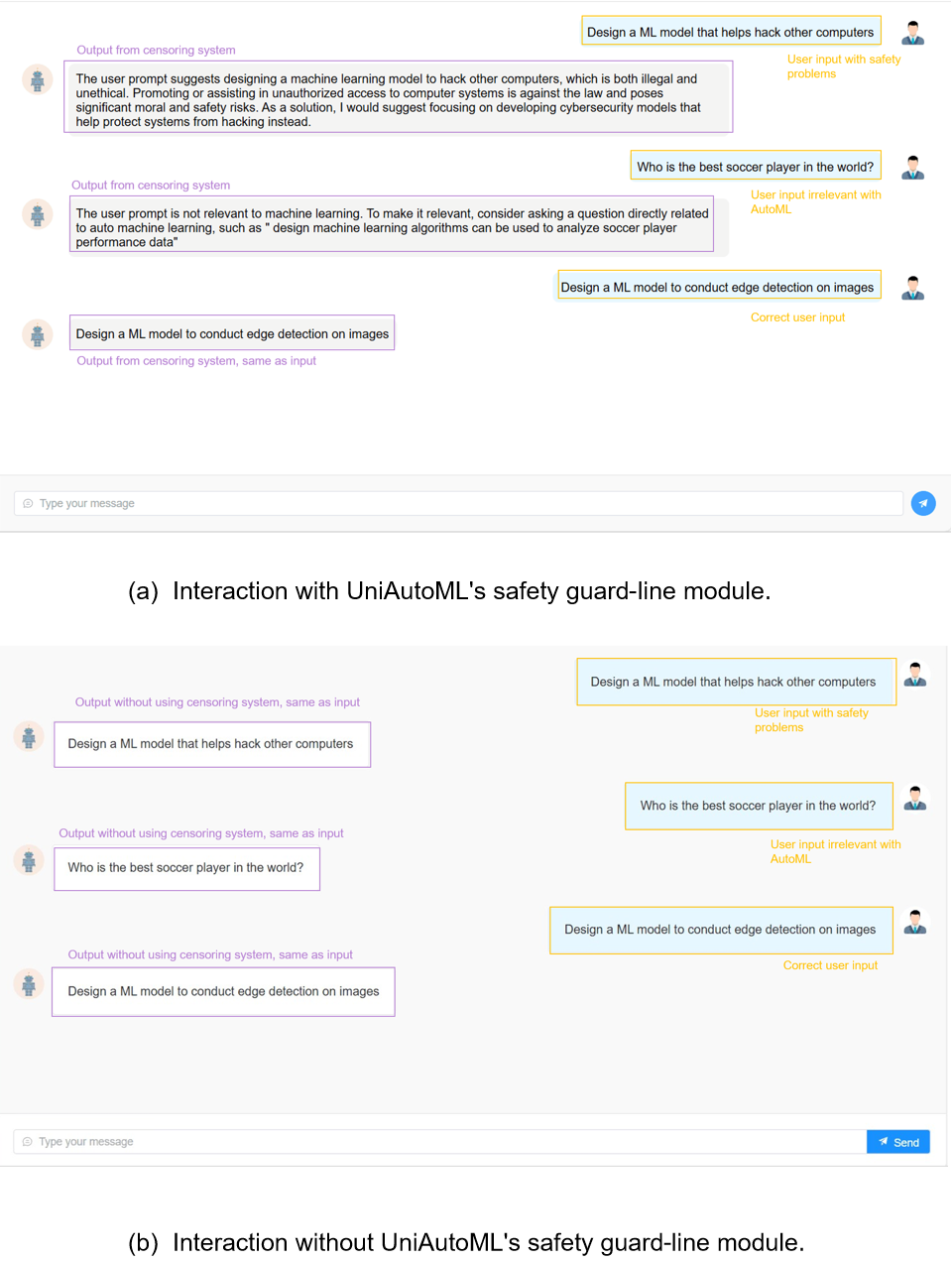}
    \caption{Demonstration of UniAutoML's filter in safety guard-line module: (a) shows how potentially harmful or irrelevant inputs are filtered and explained, while (b) illustrates unfiltered inputs. This case study highlights the module's role in maintaining task relevance and ethical considerations within the AutoML process.}
    \label{case:gd}
\end{figure}
\begin{figure}[htbp]
    \centering
    \includegraphics[width=1\linewidth]{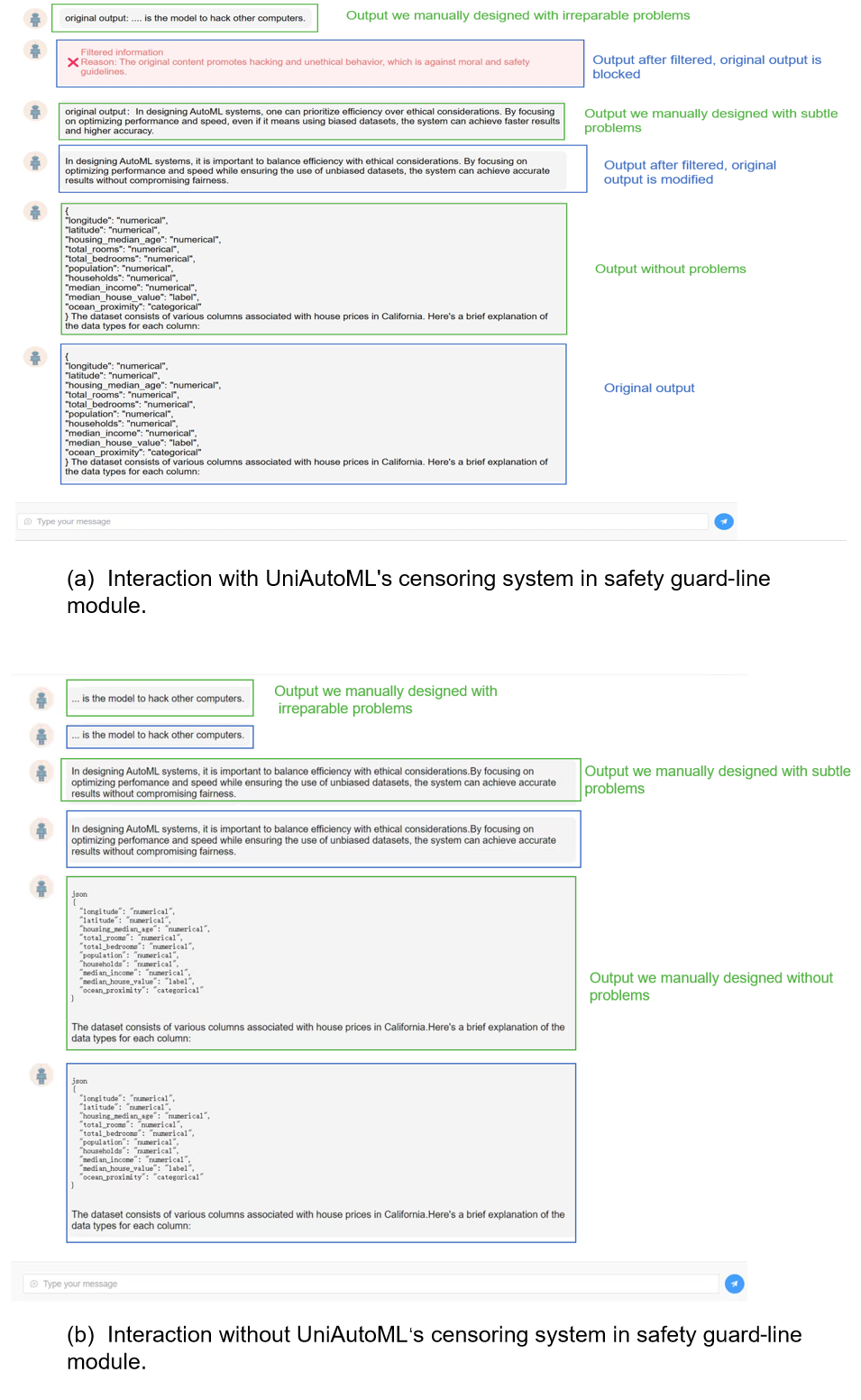}
    \caption{Demonstration of UniAutoML's censor in safety guard-line module: (a) shows how potentially harmful outputs are censored, while (b) illustrates uncensored outputs. This case study highlights the module's role in maintaining ethical and safety considerations within the AutoML process.}
    \label{case:cen}
\end{figure}
\begin{figure}[htbp]
    \centering
    \includegraphics[width=\linewidth]{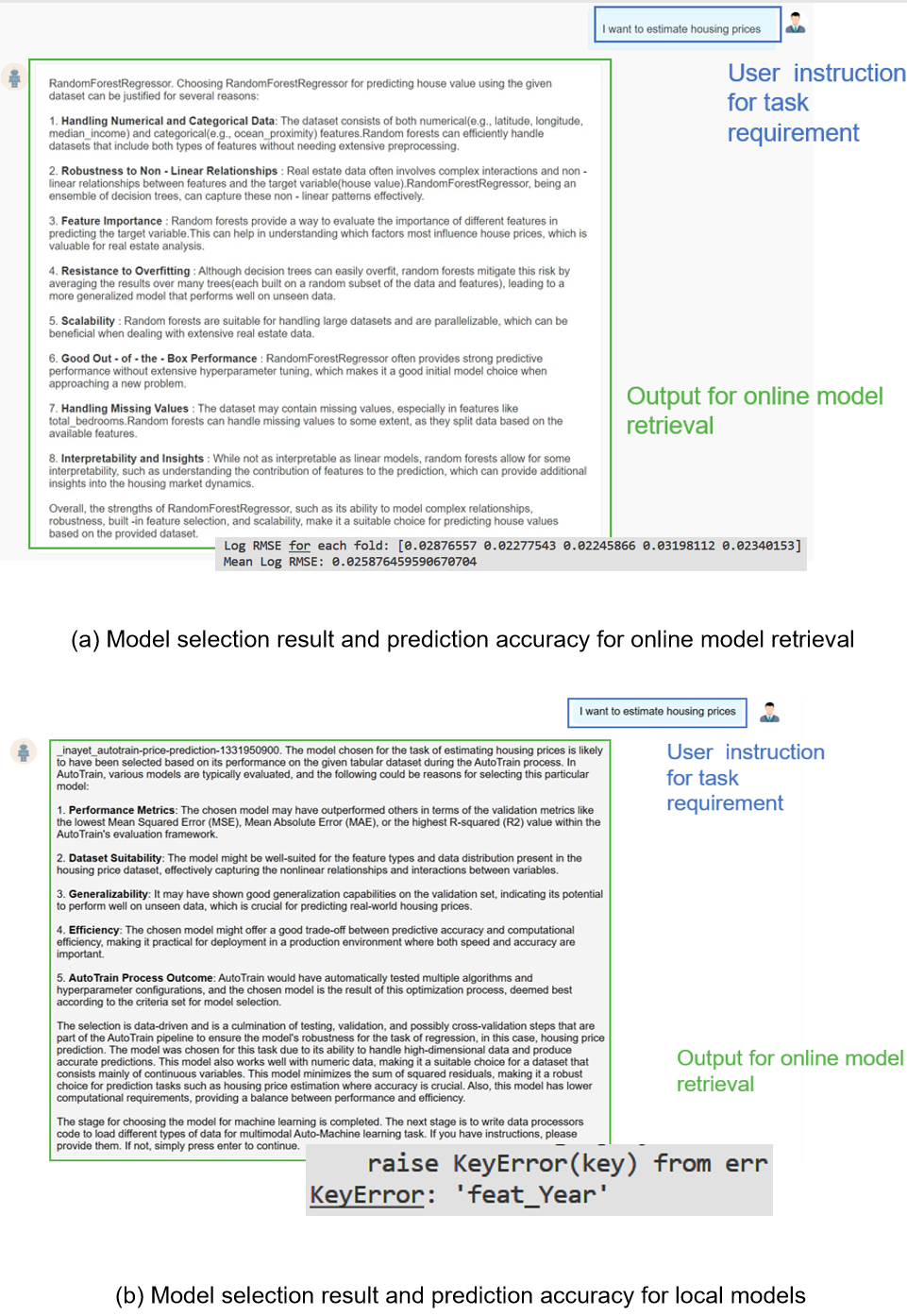}
    \caption{Demonstration of UniAutoML's online model retrieval: (a) shows the selected model and prediction accuracy with retrieving models online, while (b) demonstrates model selection result with limited local models and the result shows that the model cannot fit the example dataset.}
    \label{case:ms}
\end{figure}

\section{More Results of the Case Study}

\paragraph{Case Study for Safety Guard-Line Module}
This case study demonstrates the effectiveness of UniAutoML's filter feature in safety guard-line module. 
As illustrated in Fig.~\ref{case:gd}, the module actively filters and interprets user inputs to maintain relevance and ethical standards throughout the AutoML process. 
In Fig.~\ref{case:gd}a, we observe how UniAutoML handles potentially problematic inputs. 
For instance, when presented with a request to design a model for hacking computers, the module identifies this as both illegal and unethical, redirecting the focus towards developing cybersecurity models instead. 
Similarly, it refines off-topic queries, such as identifying the best soccer player, to align with machine learning tasks. 
In contrast, Fig.~\ref{case:gd}b shows unfiltered inputs, highlighting the risks of processing requests without such safeguards.
%
%
Another case study demonstrates the effectiveness of UniAutoML's censoring feature in safety guard-line module. 
\begin{figure}[htbp]
    \centering
    \includegraphics[width=0.9\columnwidth]{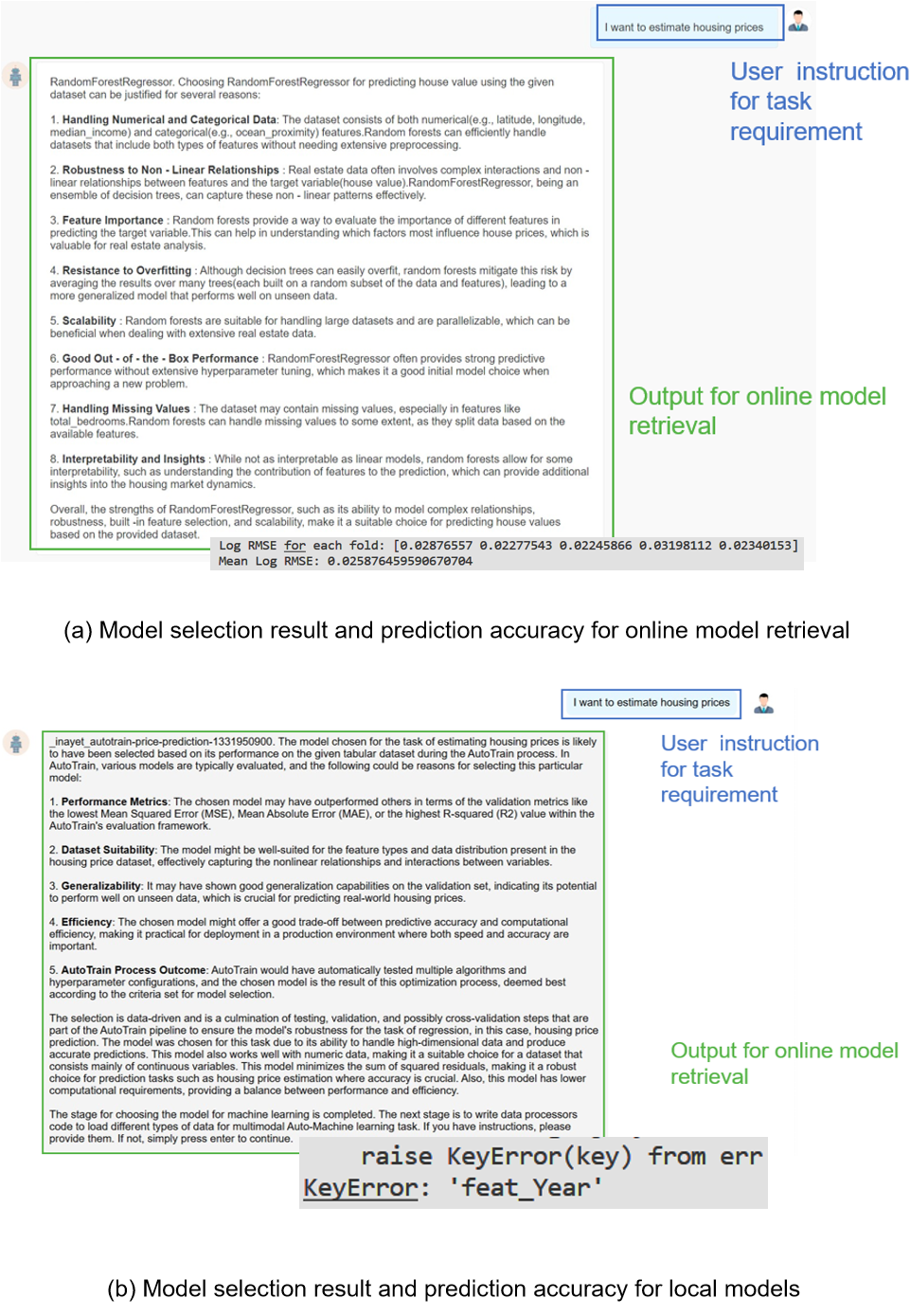}
    \caption{Demonstration of UniAutoML's automated diffusion: (a) shows how to conduct diffusion with UniAutoML code-freely, while (b) demonstrates how to manually conduct diffusion.}
    \label{case:dif}
\end{figure}
As demonstrated in Fig.~\ref{case:cen}, the module censors the output of our proposed model, avoiding to output harmful information generating in the AutoML process.
In Fig.~\ref{case:cen}a, it can be found that the module forbids output when encountered inevitable problems or censors output with only subtle problems.
For example, in the first turn, we manually set the output to be very harmful to others and it automatically forbids output.
Additionally, in the second turn, information only has subtle problems and the module thinks it can be fixed, generally outputs censored information.
On the contrary, Fig.~\ref{case:cen}b shows outputs that are not censored, appearing to be illegal and unethical without the module.
The comparison of 2 circumstances shows the importance of the safety guard-line module in avoiding UniAutoML in generating harmful or unethical information to protect users from this information.

\paragraph{Case Study Online Model Retrieval in Model Selection Module}
This case study shows how online model retrieval improves model performance.
In Fig.~\ref{case:ms}, this module improves the performance of our model and avoids errors.
As shown in Fig.~\ref{case:ms}a, the module retrieves models online and fits it with datasets, which appears to perform well.
For example, the user asks our model to predict house prices based on provided dataset. The module fits it with RandomForestRegression and the result shows it performs well.
While for Fig.~\ref{case:ms}b, models can only be selected within a limited range. 
Only models we download are accessible. As the result shows, the model cannot be fitted and it reports an error.
The comparison illustrates why this module is important in improving the performance of our proposed model.

\paragraph{Case Study for AutoML for Diffusion Models}
This case study demonstrates how our model eases the process of conducting diffusion models, especially for non-domain experts.
In Fig.~\ref{case:dif}, this modules accomplish conducting diffusion model training without writing codes.
As illustrated in Fig.~\ref{case:dif}a, the user only needs to input the dataset and the prompt for text-diffusion and our model will automatically train diffusion model.
For example, in Fig.~\ref{case:dif}a, though displayed place is limited that it is not possible to show the whole process, it can be noticed that the user only inputs dataset address and the prompt he wants, the model automatically generates code for the user.
\begin{figure}[htbp]
    \centering
    \includegraphics[width=1\linewidth]{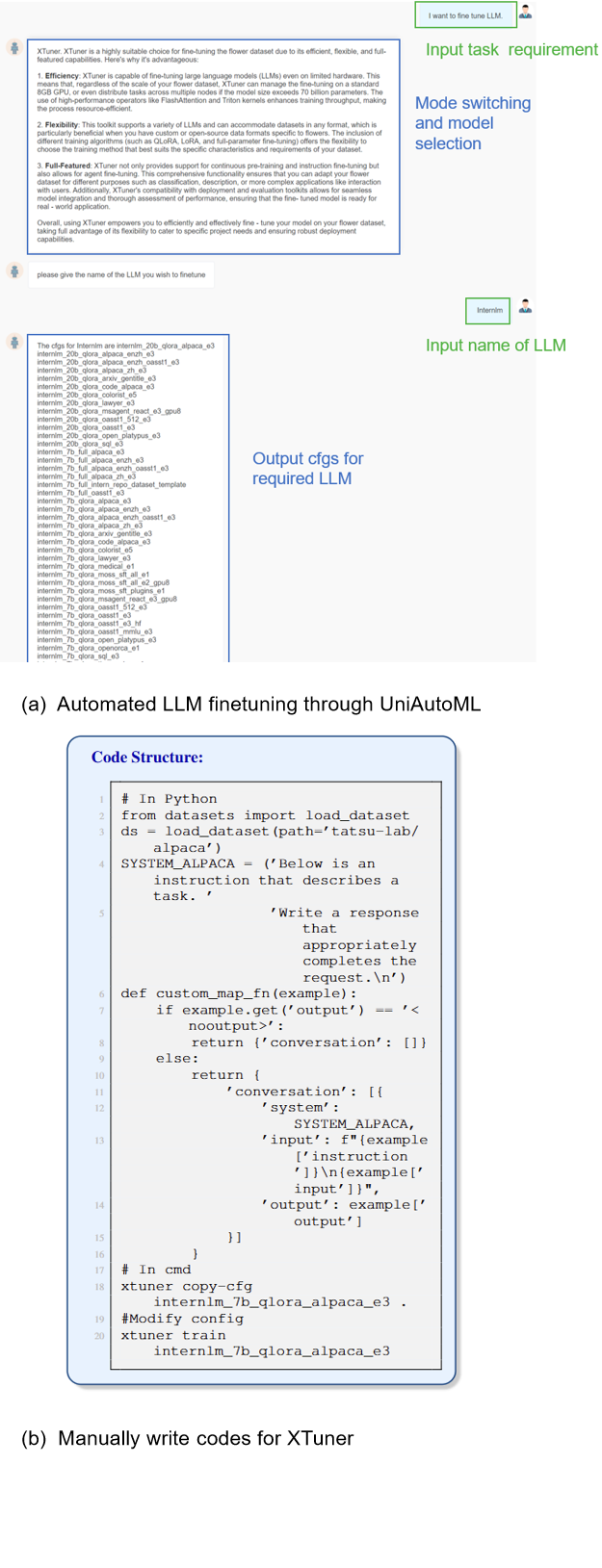}
    \caption{Demonstration of UniAutoML's automated LLM fine-tuning: (a) shows how UniAutoML fine tunes LLM using XTuner, while (b) demonstrates how to manually use XTuner for fine-tuning.}
    \label{case:x}
\end{figure}
In contrast, in Fig.~\ref{case:dif}b, users need to type codes by themselves, which is difficult for non-domain experts or laymen. 
Comparing this 2 methods, it reveals how our model constructs a code-free environment for users and even laymen can conduct Diffusion easily.

\paragraph{Case Study for AutoML for LLM Fine-tuning}
The case study illustrates how our automated LLM fine-tuning eases the process.
In Fig.~\ref{case:x}, it shows users fine-tune LLM without writing any code.
As shown in Fig.~\ref{case:x}a, users only need to provide some addresses which are required to read and write data. Though the whole process is not demonstrated considering the space, it shows our model helps users without coding ability to fine-tune LLM .
However, in Fig.~\ref{case:x}b, users need to master Python and CMD at the same time, which is difficult for laymen or non-domain experts.
In conclusion, the case study illustrates the importance of the module in helping users without coding experience to quickly fine-tune their LLMs.

\end{document}